\documentclass[10pt,twocolumn,letterpaper]{article}

\usepackage{iccv}
\usepackage{times}
\usepackage{epsfig}
\usepackage{graphicx}
\usepackage{amsmath}
\usepackage{amssymb}
\usepackage{xcolor}
\usepackage{algorithm}
\usepackage{multirow}
\usepackage{multicol}
\usepackage{subfigure}
\usepackage{threeparttable}
\usepackage{parcolumns}
\usepackage{array}
\usepackage{makecell}
\usepackage{multirow}

\usepackage[T1]{fontenc}    
\usepackage{hyperref}       
\usepackage{url}            
\usepackage{booktabs}       
\usepackage{amsfonts}       
\usepackage{nicefrac}       
\usepackage{microtype}      
\usepackage{graphicx}
\usepackage{listings}
\usepackage{color}
\usepackage{parcolumns}
\usepackage{array}
\usepackage{makecell}
\usepackage{multirow}
\usepackage{amssymb}
\usepackage{pifont}
\usepackage{graphics}
\usepackage{subfigure}
\usepackage{wrapfig}
\usepackage{algorithm}
\usepackage{algorithmic}
\usepackage{threeparttable}
\usepackage{xcolor}
\hypersetup{
colorlinks = true,
urlcolor=blue,
}

\usepackage{url}

\iccvfinalcopy 

\def\iccvPaperID{10689} 

\ificcvfinal\fi

\begin{document}

\title{FedCV: A Federated Learning Framework for Diverse Computer Vision Tasks}

\author{Chaoyang He$^1$, Alay Dilipbhai Shah$^1$, Zhenheng Tang$^2$, Di Fan$^1$ \\ Adarshan Naiynar Sivashunmugam$^1$, Keerti Bhogaraju$^1$, Mita Shimpi$^1$, Li Shen$^3$, Xiaowen Chu$^2$, \\ Mahdi Soltanolkotabi$^1$, Salman Avestimehr$^1$ \\
University of Southern California$^1$, Hong Kong Baptist University$^2$, Tencent AI Lab$^3$ \\
\textit{\{chaoyang.he, avestime\}@usc.edu}\\
}



\maketitle
\ificcvfinal\thispagestyle{empty}\fi

\begin{abstract}
   Federated Learning (FL) is a distributed learning paradigm that can learn a global or personalized model from decentralized datasets on edge devices. However, in the computer vision domain, model performance in FL is far behind centralized training due to the lack of exploration in diverse tasks with a unified FL framework.
   FL has rarely been demonstrated effectively in advanced computer vision tasks such as object detection and image segmentation. 
   To bridge the gap and facilitate the development of FL for computer vision tasks, in this work, we propose a federated learning library and benchmarking framework, named \texttt{FedCV}, to evaluate FL on the three most representative computer vision tasks: image classification, image segmentation, and object detection. 
   We provide non-I.I.D. benchmarking datasets, models, and various reference FL algorithms. Our benchmark study suggests that there are multiple challenges that deserve future exploration:  centralized training tricks may not be directly applied to FL; the non-I.I.D. dataset actually downgrades the model accuracy to some degree in different tasks; improving the system efficiency of federated training is challenging given the huge number of parameters and the per-client memory cost. We believe that such a library and benchmark, along with comparable evaluation settings, is necessary to make meaningful progress in FL on computer vision tasks. \texttt{FedCV} is publicly available: \url{https://github.com/FedML-AI/FedCV}. 
\end{abstract}
\section{Introduction}\label{introduction}
FL has the potential to rescue many interesting computer vision (CV) applications which centralized training cannot handle due to various issues such as privacy concerns (e.g. in medical settings), data transfer and maintenance costs (most notably in video analytic) \cite{zhang2021federated,Zhang2021FederatedLF}, or sensitivity of proprietary data (e.g. facial recognition) \cite{kairouz2019advances}.
In essence, FL is an art of trade-offs among many optimization objectives \cite{wang2021field,ezzeldin2021fairfed}, including improving model accuracy and personalization \cite{yu2020salvaging,fallah2020personalized,mansour2020three,singhal2021federated,huang2021personalized,hanzely2021personalized,dinh2020personalized,he2021ssfl,he2021fedgraphnn}, system efficiency (communication and computation) \cite{He2020GroupKT,lee2021layer,he2021pipetransformer,yuan2021mest,liang2021omnilytics,he2021spreadgnn}, robustness to attacks \cite{wang2019beyond,bhagoji2019analyzing,fung2018mitigating,bagdasaryan2020backdoor,wei2020framework,chen2020backdoor,sun2019can,enthoven2020overview,chen2020backdoor,BREA2020,SauravByzantine2020}, and privacy \cite{bonawitz2016practical,geyer2017differentially,orekondy2018gradient,ryffel2018generic,melis2019exploiting,truex2019hybrid,triastcyn2019federated,so2020turbo,xu2019hybridalpha,triastcyn2020federated}. There has been steady progress in FL algorithmic research to achieve these goals. 

\begin{figure}[t!] 
    \vspace{-0.4cm}
    \centering
    \includegraphics[width=0.8\linewidth]{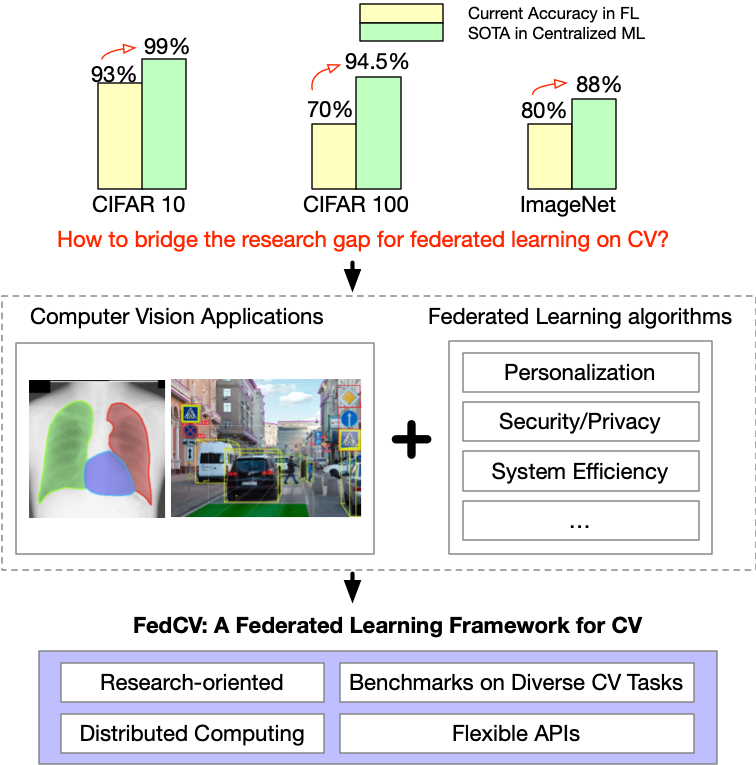} 
    \caption{Our philosophy of federated learning on computer vision: connecting the algorithmic FL research and CV application-drive research with an unified research framework.} 
    \label{fig:philosophy} 
\end{figure}

However, the research gap between computer vision (CV) \cite{He2020MiLeNASEN,huang2021lightweight} and federated learning (FL) is large. 
First, research in the FL community focuses almost exclusively on distributed optimization methods with small-scale datasets and models in image classification (see Table \ref{tab:stat-datasets} in the Appendix), while the research trends in CV focus more on large-scale supervised/self-supervised pre-training \cite{chen2020exploring} with efficient CNN \cite{efficientnet} or Transformer models \cite{dosovitskiy2020image}, which largely improves the performance of classification tasks on ImageNet and various downstream tasks such as object detection and image segmentation.
Due to the lack of exploration in diverse tasks, model performance in FL is far behind that of centralized training. 

Second, CV model training normally requires large-scale computing research in a distributed computing environment, but current FL algorithms are mostly published as standalone simulations, which further enlarges the research gap (e.g., the recently released FedVision library \cite{liu2020fedvision} only contains object detection and single GPU training).

Third, the efficacy of proposed FL algorithms on diverse CV tasks is still vague. When combined with multiple optimization objectives such as privacy, security, and fairness, the problem becomes even more challenging. Currently, only image classification in small-scale datasets and models has been evaluated in these algorithms (see Table \ref{tab:stat-datasets} in the Appendix). Researchers may attempt to solve a specific problem in realistic CV tasks by designing new algorithms, but the current research community lacks such a library to connect diverse CV tasks with algorithmic exploration.

Due to these obstacles, there is an urgent need to bridge the gap between pure algorithmic research and CV application-driven research. Our philosophy to do so can be illustrated in Figure \ref{fig:philosophy}. Specifically, we design a unified federated learning library, named \texttt{FedCV}, to connect various FL algorithms with multiple important CV tasks, including image segmentation and object detection. Under this framework, the benchmark suite is provided to assist further research exploration and fair comparison. To approach a realistic federated dataset, we provide methods to partition the dataset into non-identical and independent distribution. Model-wise, we believe the best solution for CV is to improve pre-training for SOTA models with efficient federated learning methods, which requires us to design efficient and effective task-specific models with pre-trained models as the backbone in the FL setting. To reduce the learning curve and engineering burden for CV researchers, we provide various representative FL algorithms as one line, easy-to-use APIs. Most importantly, these APIs provide distributed computing paradigm, which is essential to accelerating the federated training of most CV models. Moreover, we also make the framework flexible in exploring algorithms with new protocols of distributed computing, such as customizing the exchange information among clients and defining specialized training procedures.

To demonstrate the ability of our framework and provide benchmarking experimental results, we run experiments in three computer visions: image classification, image segmentation, and object detection. Our benchmark study suggests that there are multiple challenges that deserve future exploration: 
many deep learning training tricks may not be directly applied to FL; 
the non-IID dataset actually downgrades the model accuracy to some degree in different tasks; 
improving the system efficiency of federated training is challenging given the huge number of parameters and the per-client memory cost.
We hope \texttt{FedCV} will serve as an easy-to-use platform for researchers to explore diverse research topics at the intersection of computer vision and federated learning, such as improving models, systems, or federated optimization methods.

\section{Related Works}\label{relatedwork}
\cite{hsu2020} is the first work that applies federated learning to a real-world image dataset, Google Landmark \cite{gld}, which has now become the standard image dataset for federated learning research. \cite{chang2020, li2020, zhu2019} apply federated learning on medical image segmentation tasks, which aims at solving the issue in which the training data may not be available at a single medical institution due to data privacy regulations. In the object detection task, \cite{yu2019federated} proposes a KL divergence method to mitigate model accuracy loss due to non-I.I.D. Our work is closely related to FedVision \cite{liu2020fedvision}, a federated learning framework for computer vision. It supports object detection in the smart city scenario using models including FastRCNN and YOLOv3. However, FedVision only supports the FedAvg algorithm and single-GPU training. Our FedCV platform provides diverse computer tasks and various FL algorithms. For federated learning in other application domains, we refer to the comprehensive vision paper \cite{kairouz2019advances}.

\section{Preliminary and Challenges}

Federated learning (FL) is a distributed learning paradigm that can leverage a scattered and isolated dataset to train a global or personalized model for each client (participant) while achieving privacy preservation, compliance with regulatory requirements, and savings on communication and storage costs for such large edge data. The most straightforward formulation is to assume all clients need to collaboratively train a global model. Formally, the objective function is as follows:
\begin{equation}\label{eq:FL}
\small
\begin{split}
\min_{\boldsymbol{W}} F(\boldsymbol{W}) \stackrel{\text { def }}{=} \min_{\boldsymbol{W}} \sum_{k=1}^{K} \frac{N^{(k)}}{N} \cdot f^{(k)}(\boldsymbol{W}) \\
\space f^{(k)}(\boldsymbol{W}) = \frac{1}{N^{(k)}} \sum_{i=1}^{N^{(k)}} \ell(\boldsymbol{W}; \boldsymbol{X}_{i}, y_{i}) 
\end{split}
\end{equation}
In computer vision, $\small \boldsymbol{W}$ can be any CNN or Transformer model (e.g., ViT). $\small f^{(k)}(\boldsymbol{W})$ is the $k$th client's local objective function that measures the local empirical risk over the heterogeneous dataset $\small \mathcal{D}^k$. $\ell$ is the loss function of the global CNN model. For the image classification task, $\ell$ is the cross-entropy loss.

\begin{algorithm}[h!]
   \caption{\texttt{FedAvg} Algorithm: A Challenge Perspective}
   \label{alg:transformation algorithm}
\begin{algorithmic}[1]
\STATE {\bfseries Initialization:} there is a number of clients in a network; \colorbox{green!20}{the client $k$ has local dataset $\small \mathcal{D}^k$}; each client's local model is initialized as $\boldsymbol{W}_0$; 
\STATE
\STATE \textbf{Server\_Executes:}
   \FOR{each round $t = 0, 1, 2, \dots$}
     \STATE $\small S_t \leftarrow$ (sample a random set of clients)
     \FOR{each client $k \in S_t$ \textbf{in parallel}}
       \STATE $\small \boldsymbol{W}_{t+1}^k \leftarrow \text{ClientUpdate}(k, \boldsymbol{W}_t)$ 
     \ENDFOR
     \STATE $\small \boldsymbol{W}_{t+1} \leftarrow \sum_{k=1}^K \frac{n_k}{n}\boldsymbol{W}_{t+1}^k$
   \ENDFOR
   \STATE
\STATE \colorbox{red!20}{\textbf{Client\_Training($\small k, \boldsymbol{W}$):}}  // \emph{Run on client $k$}
  \STATE $\small \mathcal{B} \leftarrow$ (split $\small \mathcal{D}^k$ into batches)
  \FOR{each local epoch $i$ with $i=1,2,\cdots$}
    \FOR{batch $\small b \in \mathcal{B}$}
      \STATE $\small \boldsymbol{W} \leftarrow \boldsymbol{W} - \eta \nabla_{\boldsymbol{W}}F(\boldsymbol{W}; b)$
    \ENDFOR
 \ENDFOR
 \STATE \colorbox{blue!30}{return $\small \boldsymbol{W}$ to server}
\end{algorithmic}
\label{alg:autopipe}
\end{algorithm}

To solve this federated optimization problem, FedAvg is the first federated optimization algorithm to propose the concept of FL. To better understand the challenges of FL on CV, we rewrite its optimization process in Algorithm 1 with annotations. As we can see, there are several clear characteristics that distinguish FL from conventional distributed training in a sealed data center:

\textit{\colorbox{green!20}{1. Data heterogeneity and label deficiency at the edge.}} In conventional distributed training, centralized datasets are evenly distributed into multiple compute nodes for parallel computing, while in FL, the data is generated at the edge in a property of non-identical and independent distribution (non-I.I.D.). For example, in the CV scenario, smartphone users generate images or videos with distinct resolutions, qualities, and contents due to differences in their hardware and user behaviors. In addition, incentivizing users to label their private image and video data is challenging due to privacy concerns. 

\textit{\colorbox{red!20}{2. System constraints and heterogeneity.}} Training large DNN models at the edge is extremely challenging even when using the most powerful edge devices. In terms of memory, edge training requires significantly more memory than the edge inference requires. The bandwidth for edge devices is smaller than that of distributed training in the data center environment (InfiniBand can be used); the edge devices normally do not have GPU accelerators. What is even worse is that these system abilities are heterogeneous due to diverse hardware configurations.

\colorbox{blue!30}{3. Robustness and Privacy.} Since federated training is not in a sealed data center environment, as is the traditional distributed training, it is easier to manipulate the data and model poisoning attacks. Therefore, making the training algorithm robust against attacks is also an important research direction in FL. In addition, although privacy preservation is one of the main goals, researchers also demonstrate that the exchanged gradient between the client and the server may, to some degree, lead to privacy leaks. More privacy-preserving techniques must be evaluated on various computer vision applications \cite{yang2021lightsecagg}.

\section{\texttt{FedCV} Design}
\label{method}

To solve these challenges in diverse CV tasks, we need a flexible and efficient distributed training framework with easy-to-use APIs, benchmark datasets and models, and reference implementations for various FL algorithms.

\begin{figure}[h!]
\centering
{\includegraphics[width=0.9\linewidth]{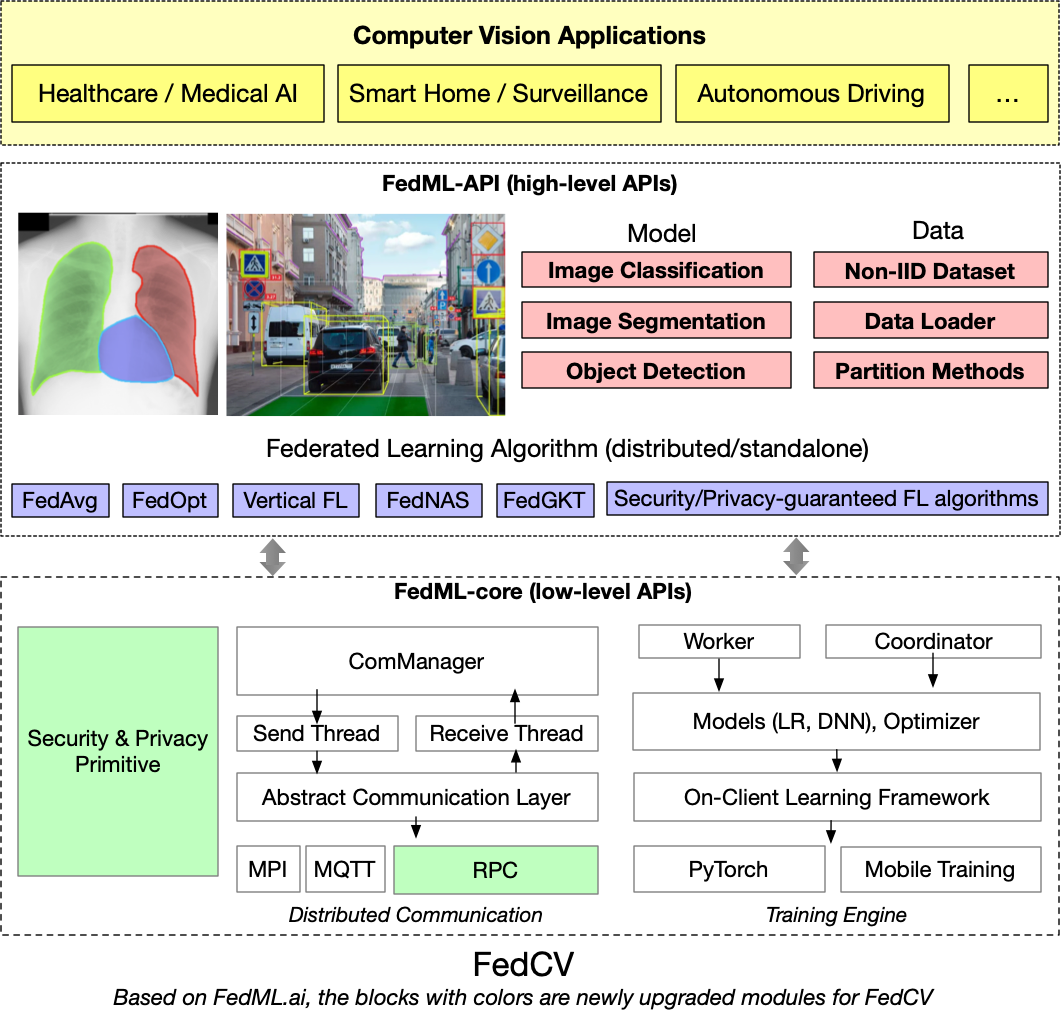}}
\caption{Overview of \texttt{FedCV} System Architecture Design}
\label{fig:architecture}
\end{figure}


To bridge the gap between CV and FL research, we have designed an open-source federated learning system for computer vision, named \texttt{FedCV}. \texttt{FedCV} is built based on the \texttt{FedML} research library \cite{He2020FedMLAR}, which is a widely used FL library that only support image classification, ResNet and simple CNN models. The system architecture of \texttt{FedCV} is illustrated in Figure~\ref{fig:architecture}. To distinguish FedCV from FedML, we color-code the modules specific to \texttt{FedCV}. \texttt{FedCV} makes the following contributions:  

\textit{Benchmark Suite for Diverse CV Tasks:} \texttt{FedCV} supports three computer vision tasks: image classification, image segmentation, and object detection. Related datasets and data loaders are provided. Users can either reuse our data distribution or manipulate the non-I.I.D. by setting hyper-parameters. Models are curated for benchmark evaluation. More details of the benchmark suite are given in Section \ref{benchmarksuite}.

\textit{Reference Implementation for Representative FL Algorithms:} Currently, \texttt{FedCV} includes the standard implementations of multiple state of the art FL algorithms: Federated Averaging (FedAvg) \cite{mcmahan2017communication}, FedOpt (server Adam) \cite{reddi2020adaptive}, FedNova (client optimizer) \cite{wang2020tackling}, FedProx \cite{Sahu2018OnTC}, FedMA \cite{wang2020federated}, as well as some novel algorithms that have diverse training paradigms and network typologies, including FedGKT (efficient edge training) \cite{He2020GroupKT}, Decentralized FL \cite{he2019central}, Vertical Federated Learning (VFL) \cite{fmlyangqiang}, Split Learning \cite{gupta2018distributed, vepakomma2018split}, Federated Neural Architecture Search (FedNAS) \cite{he2020fednas}, and Turbo-Aggregate \cite{so2020turbo}. These algorithms support multi-GPU distributed training, which enables training to be completed in a reasonable amount of time. Note that most published FL optimization algorithms are based on standalone simulations, which lead to a extremely long training time. In this paper, we bridge this gap and make the CV-based FL research computationally affordable.

\textit{Easy-to-use APIs for Algorithm Customization:} With the help of the \texttt{FedML} API design, \texttt{FedCV} enables diverse networks, flexible information exchange among workers/clients, and various training procedures. We can easily implement new FL algorithms in a distributed computing environment. We defer API design details to the Appendix.

\textit{Other Functionality:} We support several development tools to simplify the research exploration. Specifically, researchers can load multiple clients into a single GPU, which scales up the client number with fewer GPUs, although there may be GPU contention among processes; in the lowest layer, \texttt{FedCV} reuses \texttt{FedML-core} APIs but further supports tensor-aware RPC (remote procedure call), which enables the communication between servers located at different data centers (e.g., different medical institutes); enhanced security and privacy primitive modules are added to support techniques such as secure aggregation in upper layers.

\section{\texttt{FedCV} Benchmark Suite: Datasets, Models, and Algorithms}
\label{benchmarksuite}


\begin{table}[h!]
 \small
  \centering
  \fontsize{10}{10}\selectfont
  \begin{threeparttable}
  \caption{Summary of benchmark suite.}
  \label{tab:benchmark_setting}
    \begin{tabular}{ccc}
    \toprule
    Task & Dataset & Model\cr
    \midrule
    \multirow{3}{*}{ \!\!\!{\small Image Classification}\!\!\!} & \multirow{2}{*}{{\small CIFAR-100}}  & {\small EfficientNet\cite{efficientnet}} \cr
    &  & {\small MobileNet\cite{mobilenet}} \cr
    & {\small GLD-23k\cite{gld}} &  {\small ViT\cite{ViT}} \cr
    \hline
    \multirow{2}{*}{\!\!\!{\small Image Segmentation}\!\!\!} & \multirow{2}{*}{ {\small PASCAL VOC\cite{BharathICCV2011}}}\!\!\!  & {\small DeeplabV3+} \cr
    & & {\small UNet\cite{U-Net}} \cr
    \hline
    \specialrule{0em}{1pt}{1pt}
    {\small Object Detection}     & {\small COCO\cite{lin2014microsoft}}                  & {\small YOLOv5\cite{YOLOv5}}  \cr
    \hline
    \hline
    \specialrule{0em}{1pt}{1pt}
     {\small FL Algorithms} & \multicolumn{2}{c}{{\small FedAvg, FedOpt ...  }} \cr
    \bottomrule
    \end{tabular}
    \end{threeparttable}
\end{table}

We summarize the benchmark suite in \texttt{FedCV} in Table \ref{tab:benchmark_setting}, and introduce such a curated list task-by-task as follows:

\paragraph{Image Classification.} The curated datasets are Google Landmarks Dataset 23k (GLD-23K) \cite{gld} and CIFAR-100 dataset \cite{cifar100} with non-I.I.D partition. GLD-23K dataset is suggested by Tensorflow Federated \cite{gldtf}, a natural federated dataset from smartphone users. For the model, we suggest EfficientNet \cite{efficientnet} and MobileNet-V3 \cite{mobilenet}, which are two lightweight CNNs.  Since the attention-based Transformer model has become a trending model in CV, we suggest Vision Transformer (ViT) \cite{ViT} (ViT-B/16) to conduct experiments. As the research progresses, we may be able to support more efficient Transformers.

\paragraph{Image Segmentation.} We use the augmented PASCAL VOC dataset with annotations from 11355 images \cite{BharathICCV2011}. These images are taken from the original PASCAL VOC 2011 dataset which contains 20 foreground object classes and one background class. For models, DeepLabV3+ \cite{DeepLabV3+} and U-Net \cite{U-Net} are supported since they are representative image segmentation models in centralized training. In our experiments, we utilize ResNet-101 and MobileNet-V2 as two backbones of DeepLabV3+ and U-Net.

\paragraph{Object Detection.} We use the COCO \cite{lin2014microsoft} dataset since it contains realistic images that include detecting and segmenting objects found in everyday life through extensive use of Amazon Mechanical Turk. We then use YOLOv5 \cite{YOLOv5}, an optimized version of YOLOv4 \cite{bochkovskiy2020yolov4} as the baseline model. It outperforms all the previous versions and approaches EfficientDet Average Precision(AP) with higher frames per second (FPS). In YOLOv5, four network models (YOLOv5s, YOLOv5m, YOLOv5l, YOLOv5x) with different network depths and widths are provided to cater to various applications. We use these four models as the pretrained models.

\begin{figure}[h!]
\small
    \centering
    \includegraphics[height=4.3cm, width=8.3cm]{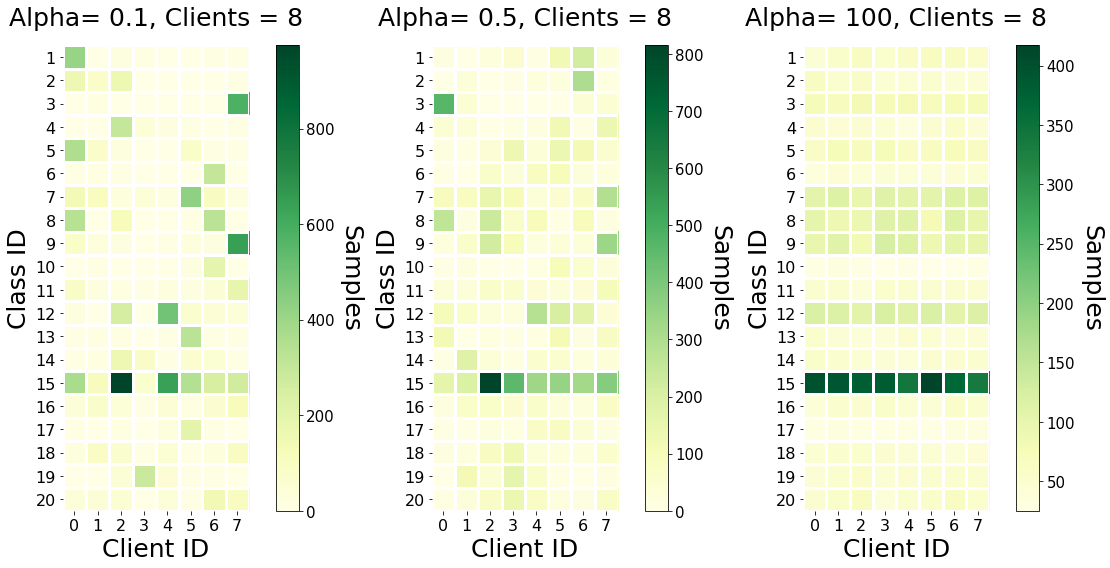}
    \caption{Non I.I.D. data distribution on augmented PASCAL VOC dataset with different $\alpha$ values. Each square represents the number of data samples of a specific class for a client.}
    \label{fig:heatmap}
\vspace{-0.8cm}
\end{figure}

\paragraph{Non-I.I.D. Preprocessing.} Our non-I.I.D. partition method is Latent Dirichlet Allocation (LDA) \cite{LDA}, which is a common practice in FL research to obtain synthetic federated datasets. As an example, we visualize the non-I.I.D. in Figure \ref{fig:heatmap}. Further details of the dataset partition can be found in the Appendix. For all datasets, we provide data downloading scripts and data loaders to simplify the data preprocessing. Note that we will update regularly to support new datasets and models.

The supported FL algorithms include FedAvg, FedOpt and many other representative algorithms. We provide a full list and description in the appendix.

\begin{figure*}[h!]
\small
\vspace{-0.2cm}
   \centering
     \!\!\!\!\!\!\!\!\subfigure[Three models on GLD-23k]{\includegraphics[width=0.255\textwidth]{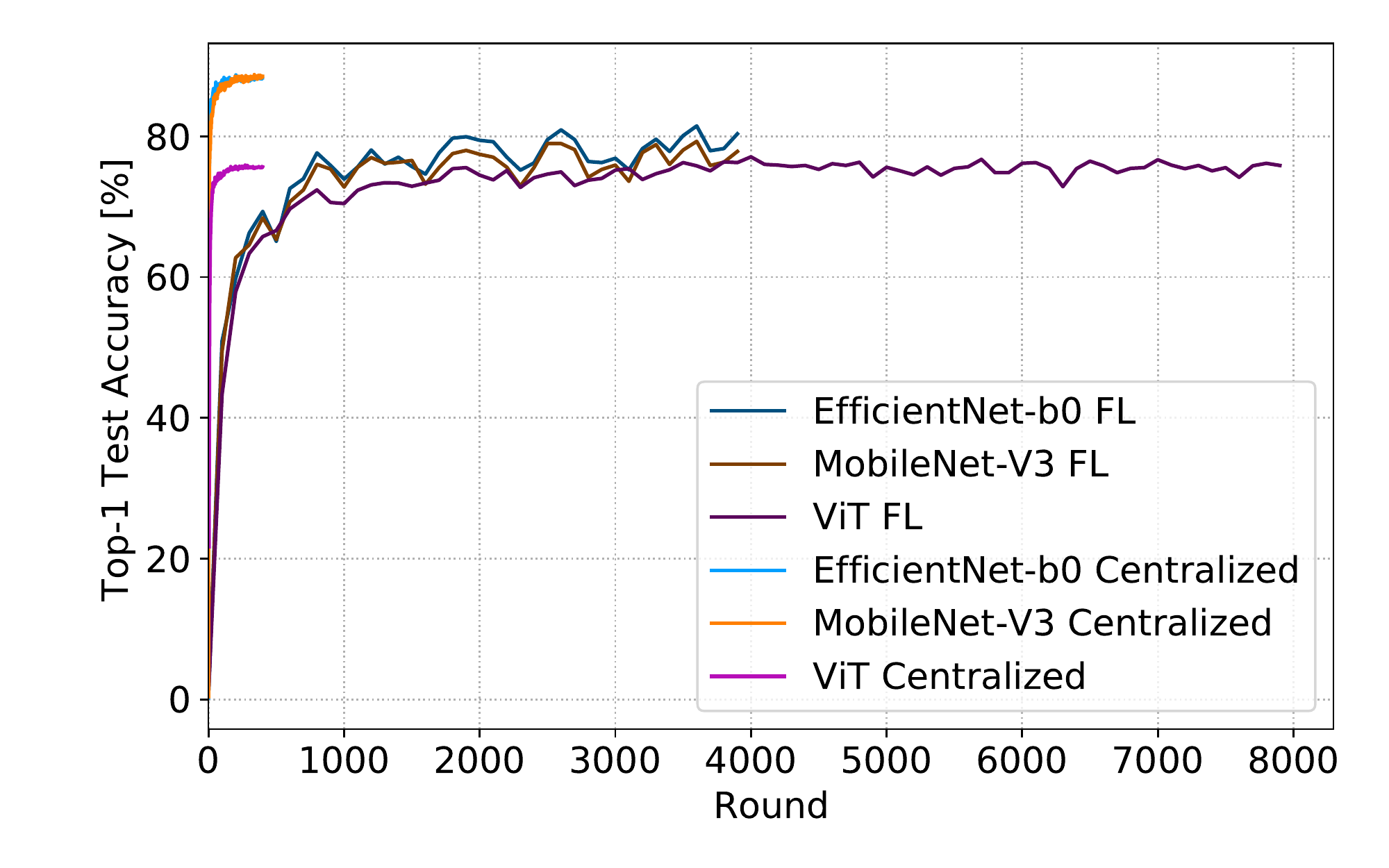}}\!\!\!\!
  \subfigure[EfficientNet on CIFAR-100]{\includegraphics[width=0.255\textwidth]{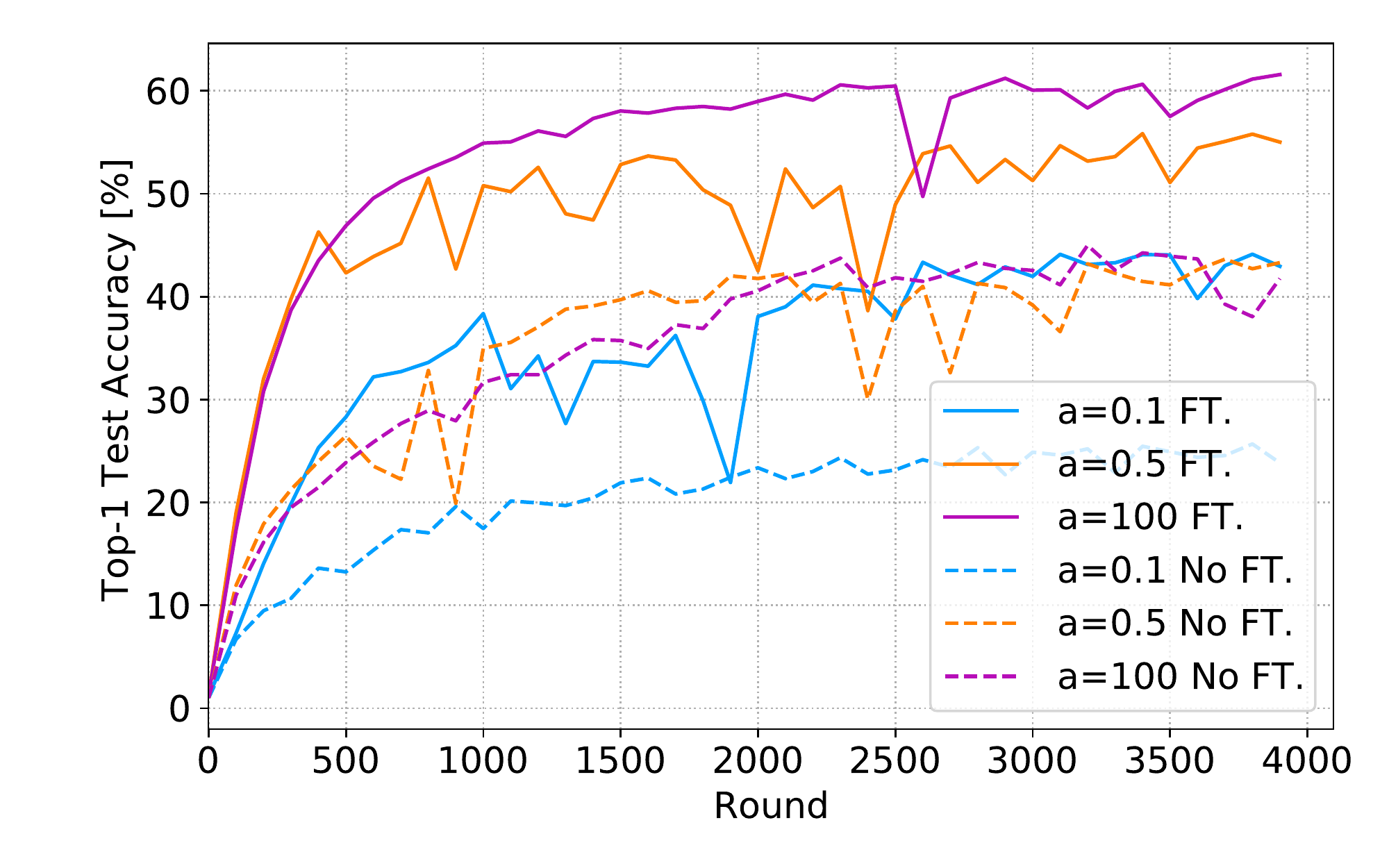}}\!\!\!\!
  \subfigure[Three models on GLD-23k]{\includegraphics[width=0.255\textwidth]{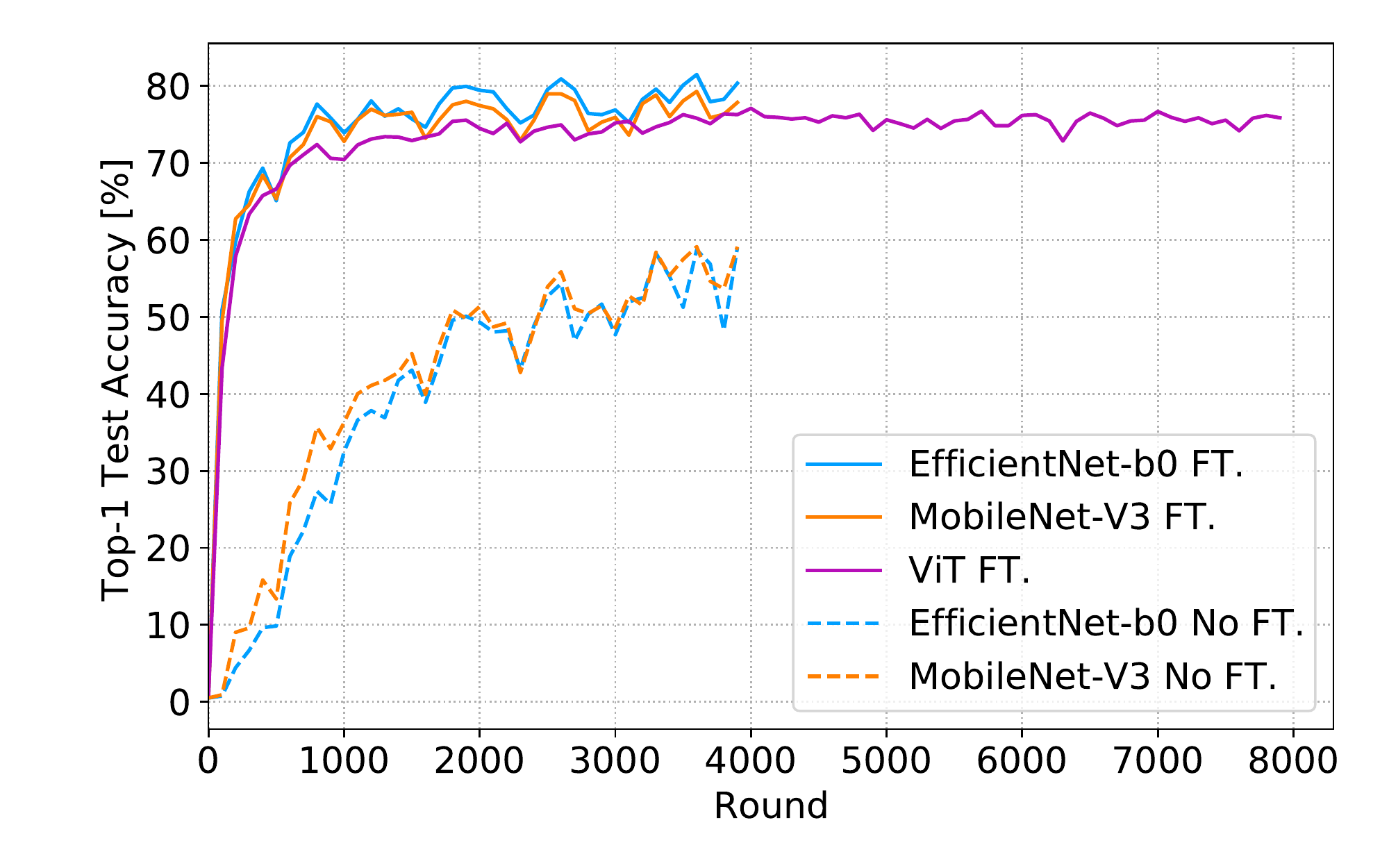}}\!\!\!\!
  \subfigure[EfficientNet on CIFAR-100]{\includegraphics[width=0.255\textwidth]{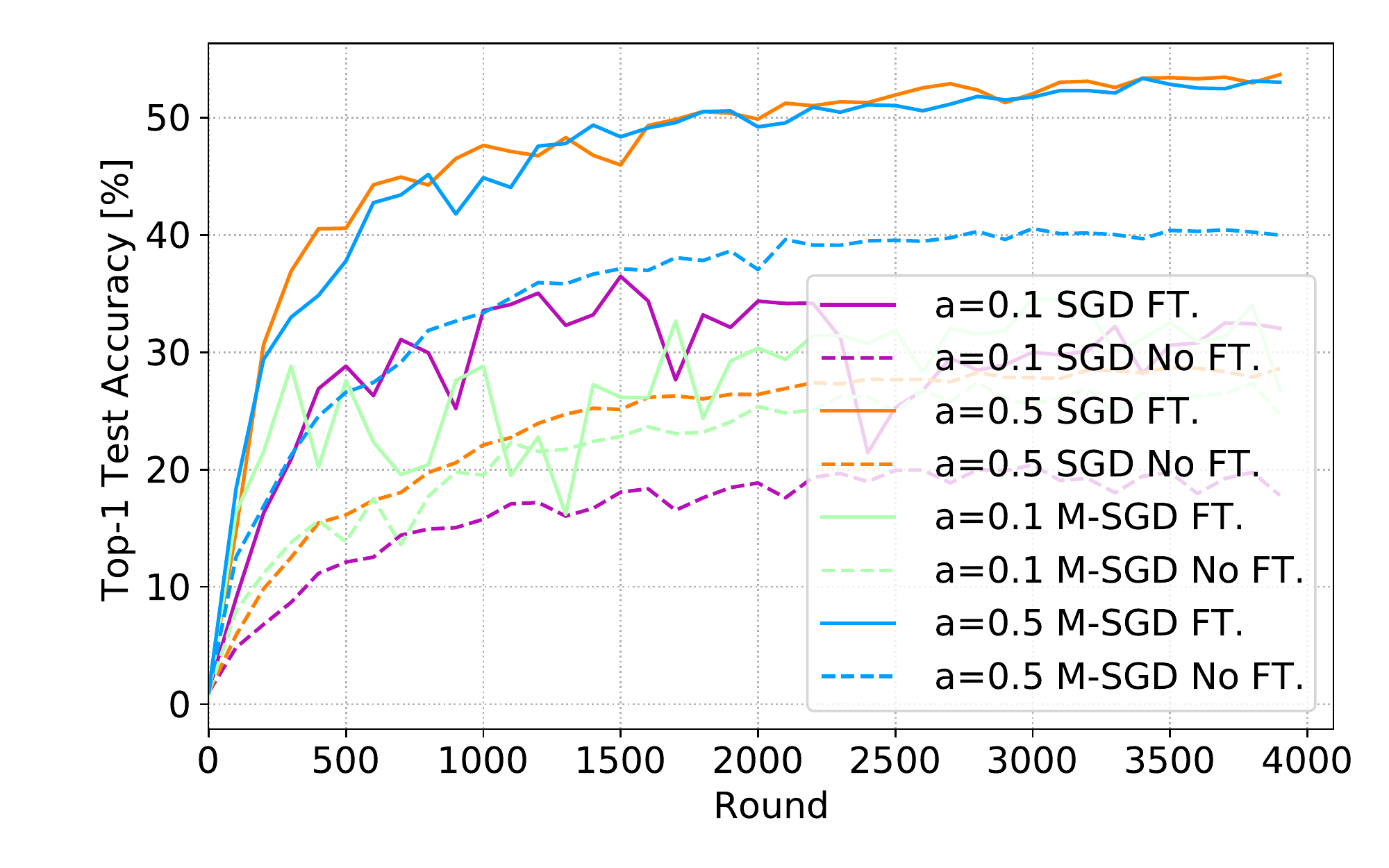}}\!\!\!\!\!
    \caption{Experiments on classification task. Figure (a): Test accuracy on GLD-23k with FedAvg and centralized training. The maximum number of epochs of centralized training is 400. The learning rate is 0.3 for EfficientNet and MobileNet of centralzed training, 0.03 for ViT of centralized traning, and 0.1 for all three models of FedAvg. Here the learning scheduler is not used for FedAvg.
    Figure (b): Test accuracy of FedAvg with EfficientNet on CIFAR-100 with different Non-IID degree. Hyper-parameters of this figure are set as Table \ref{tab:summary_of_cifar100_efficientnet} in the appendix. Here, FT. means fine-tuning, i.e. loading a pretrained model and doing FedAvg on this model.
    Figure (c): Test accuracy of FedAvg with EfficientNet, MobileNet and ViT on GLD-23K, with/without fine-tuning. Hyper-parameters of this figure can be found in Tables \ref{tab:summary_of_gld23k_efficientnet}, \ref{tab:summary_of_gld23k_mobilenet} and \ref{tab:summary_of_gld23k_vit} in appendix.
    Figure (d): Test accuracy of FedAvg with EfficientNet on CIFAR-100, with/without fine-tuning, SGD or momentum SGD. Hyper-parameters of this figure are set as Table \ref{tab:summary_of_cifar100_efficientnet} and Table \ref{tab:summary_of_cifar100_mobilenet} in appendix. Here, M-SGD means using local SGD with momentum.
    }
    \label{fig:Image_classification}
\end{figure*}

\section{Experiments}
In this section, we present the experimental results on image classification, image segmentation, and objective detection tasks on varying deep learning models and datasets.

\subsection{Image Classification}

\subsubsection{Implementation Details}

For image classification, the client number per round is 10. 
All experiments are conducted on a computing cluster with GTX 2080Ti. Each client has one GPU and the communication bandwidth is 10 Gbps. 
We conduct extensive experiments with EfficientNet \cite{efficientnet}, MobileNet V3 \cite{mobilenet} and ViT \cite{ViT} on CIFAR-100 \cite{cifar100}, and CLD-23K \cite{gld}\cite{gldtf} datasets. 
The hyper-parameter settings are listed in the Appendix.

\subsubsection{Experimental Results}

The main experimental results are presented in table \ref{tab:summary_of_all_results}. Below, we provide detailed comparisons of the implemented classification models on the proposed FedCV platform.

\begin{table}[h!]
  \centering
  \fontsize{8}{8}\selectfont
  \begin{threeparttable}
    \begin{tabular}{ccccc}
    \toprule
Dataset                    & Model                         & Partition & LR    & Acc     \\ 
\hline
\multirow{8}{*}{CIFAR-100} & \multirow{4}{*}{EfficientNet} & Cent.     & 0.01  & 0.6058  \\ 
                           &                               & a=0.1     & 0.003 & 0.4295  \\
                           &                               & a=0.5     & 0.01  & 0.5502  \\ 
                           &                               & a=100.0   & 0.003 & 0.6158  \\ 
\cline{2-5}
                           & \multirow{4}{*}{MobileNet V3} & Cent.     & 0.01  & 0.5785  \\ 
                           &                               & a=0.1     & 0.003 & 0.4276  \\ 
                           &                               & a=0.5     & 0.01  & 0.4691  \\ 
                           &                               & a=100.0   & 0.003 & 0.5203  \\ 
\hline
\multirow{6}{*}{GLD-23k}   & \multirow{2}{*}{EfficientNet} & Cent.     & 0.3   & 0.8826  \\ 
                           &                               & Non-IID   & 0.1   & 0.8035  \\ 
\cline{2-5}
                           & \multirow{2}{*}{MobileNet V3} & Cent.     & 0.3   & 0.8851  \\ 
                           &                               & Non-IID   & 0.03  & 0.7841  \\ 
\cline{2-5}
                           & \multirow{2}{*}{ViT-B/16}     & Cent.     & 0.03  & 0.7565  \\ 
                           &                               & Non-IID   & 0.03  & 0.7611  \\
    \bottomrule
    \end{tabular}
    \end{threeparttable}
\vspace{0.1cm}
\caption{Summary of experimental results on image classification. In this table, Cent. refers to centralized training. For all experiments, we use a batch size of 256 for centralized training and $32$ for FedAvg. We use a linear learning rate scheduler with a step size of 0.97 for centralized training, but no scheduler for FedAvg. We use momentum SGD with momentum coefficient of 0.9 for all experiments. More experimental results on other settings can be found in Tables \ref{tab:summary_of_cifar100_efficientnet}, \ref{tab:summary_of_cifar100_mobilenet}, \ref{tab:summary_of_gld23k_efficientnet}, \ref{tab:summary_of_gld23k_mobilenet} and \ref{tab:summary_of_gld23k_vit} in the {\bf Appendix}.}
\label{tab:summary_of_all_results} 
\end{table}

\noindent
\textbf{GLD-23k NonIID vs. IID.}\  Figure \ref{fig:Image_classification}(a) shows that the test accuracy of centralized training with EfficientNet and MobileNet outperforms FedAvg training by ten percent. And for the ViT, the accuracy of centralized training is similar with FedAvg.

\noindent
\textbf{Impacts of different degrees of Non-IID.}\  Figure \ref{fig:Image_classification}(b) and Figure \ref{fig:Image_classification_cifar100_convergence_mobilenet} (in Appendix) show the influence of different degrees of Non-IID on the training performance of EfficientNet and MobileNetV3. Experimental results align with the results of LDA \cite{LDA}. A higher $\alpha$ (i.e., lower degree of Non-IID) causes the test accuracy to increase.

\noindent
\textbf{Fine-tuning vs. training from scratch.}\  Figure \ref{fig:Image_classification}(b), Figure \ref{fig:Image_classification_cifar100_convergence_mobilenet} in appendix, and Figure \ref{fig:Image_classification}(c) show that the performance of fine-tuning is more effective than training from scratch. 
For the convergence speed, fine-tuning can achieve a test accuracy of 60\%, nearly 20$\times$ faster than training from scratch. 
After training is completed, fine-tuning outperforms training from scratch by about 20 percent.

\noindent
\textbf{Momentum SGD vs SGD.}\  Figure \ref{fig:Image_classification} (d), and Figures \ref{fig:appendix_classification}(a)-(b) (in appendix) show that SGD with momentum cannot guarantee better performance than vanilla SGD. When using EfficientNet On CIFAR-100 dataset of $\alpha = 0.5$, momentum SGD has similar performance to SGD with fine tuning, but with a much higher test accuracy than SGD training from scratch. With $\alpha = 0.1$, the performance of momentum SGD is not significantly influenced by fine-tuning, whereas vanilla SGD can see significant improvement.


\begin{figure}[h!]
\small
    \centering
    \!\!\!\!\!\!\includegraphics[width=6.5cm]{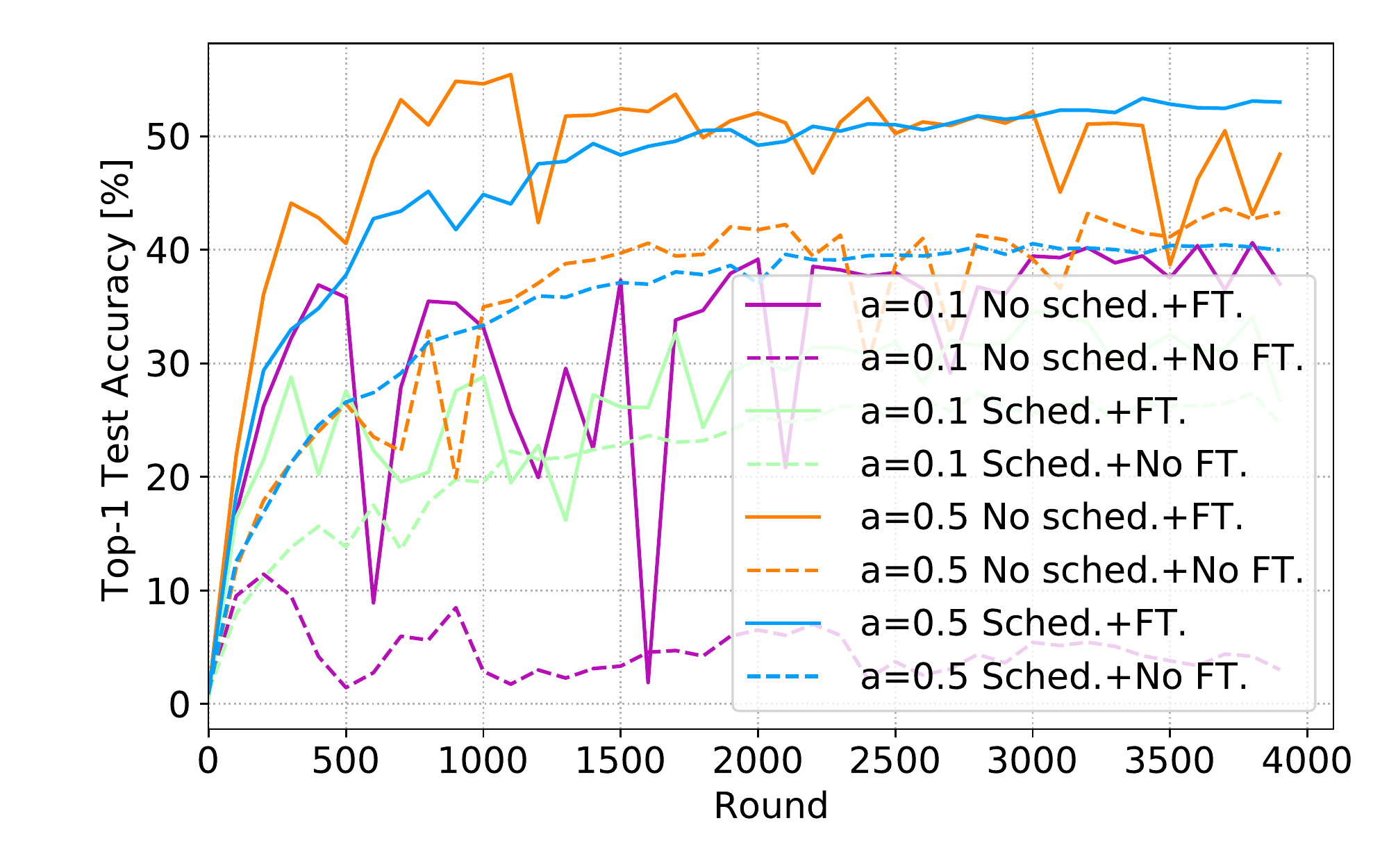}
    \caption{Test accuracy on CIFAR-100 with EfficientNet trained with momentum SGD, with/without fine-tuning and learning rate scheduler. Hyper-parameters  are set as Table \ref{tab:summary_of_cifar100_efficientnet} in appendix. Here, Sched. means using learning rate scheduler with step size of 0.99.}
    \label{fig:Image_classification_cifar100_scheduler_efficientnet}
\vspace{-0.1cm}
\end{figure}

\noindent
\textbf{Learning rate scheduler.}\  Figure \ref{fig:Image_classification_cifar100_scheduler_efficientnet}, and Figure \ref{fig:appendix_classification}(c)-(d) (in appendix) show an interesting result in which the linear learning rate decay may not improve the performance, and even leads to performance decrease. One reason may be that in the last training epochs, each client cannot converge with too small learning rate. However, learning rate decay is able to make the training process more stable. For cases where $\alpha=0.1$ and $\alpha=0.5$, four curves of linear learning rate decay are smoother than without learning rate decay. 


\begin{table}[h!]
\small
\centering
\begin{tabular}{|l|l|l|l|} 
\hline
Model      & MobileNet-V3 & EfficientNet                         & ViT-B/16                                 \\ 
\hline
Params     & 4M                                      & 3.8M                                    & 81.8M                                    \\ 
\hline
MMACs      & 2137               & 3796.4             & 16067.5             \\  \hline
Comm rounds & 4000 & 4000 & 8000 \\ \hline
Total time & 5.16h & 5.05h & 31.1h \\  \hline
Comm cost  & 0.278h                                  & 0.264h                                  & 5.68h                                    \\ \hline
\end{tabular}
\vspace{0.1cm}
\caption{Efficiency of training MobileNet V3, EfficientNet, Vit models with FedAvg. In this table, MMACs refer to the forward computation for one sample. Total time refers to the entire training time plus evaluating time; we evaluate the model per 100 communication rounds. 
For the MobileNet and EfficientNet, the number of total communication rounds is 4000, and for ViT it is 8000. 
The communication cost is theoretically calculated out. Note the actual communication time should be larger than the theoretical communication time due to the straggler problem and other overhead.}
\label{tab:summary_of_models}

\vspace{-0.2cm}
\end{table}

\noindent
\textbf{Efficiency analysis.}\  We summarize the system performance of three models in Table \ref{tab:summary_of_models}, which demonstrate that if we train a big deep learning model such as ViT in the federated setting, there exists a huge communication overhead compared with small models. Furthermore, in the real federated environment, the communication bandwidth could be even worse.


\subsection{Image Segmentation}
\subsubsection{Implementation Details}

\begin{figure*}[htb!]
\small
\vspace{-0.2cm}
  \centering
  \!\!\!\!\!\!\subfigure[Experiments with/without fine-tuning]{\includegraphics[width=9cm]{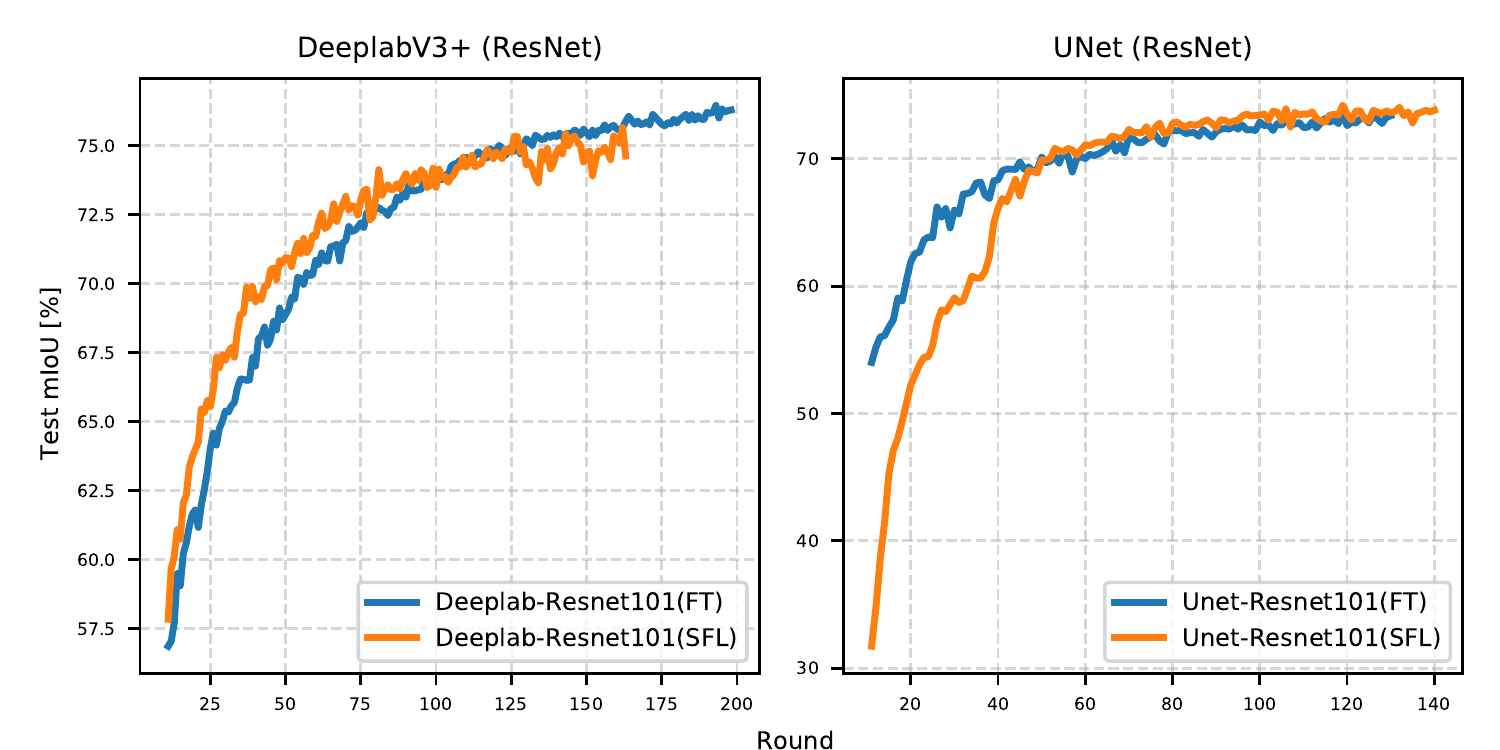}}\!\!  
  \!\!\subfigure[Experiments with varying batch sizes]{\includegraphics[width=9cm]{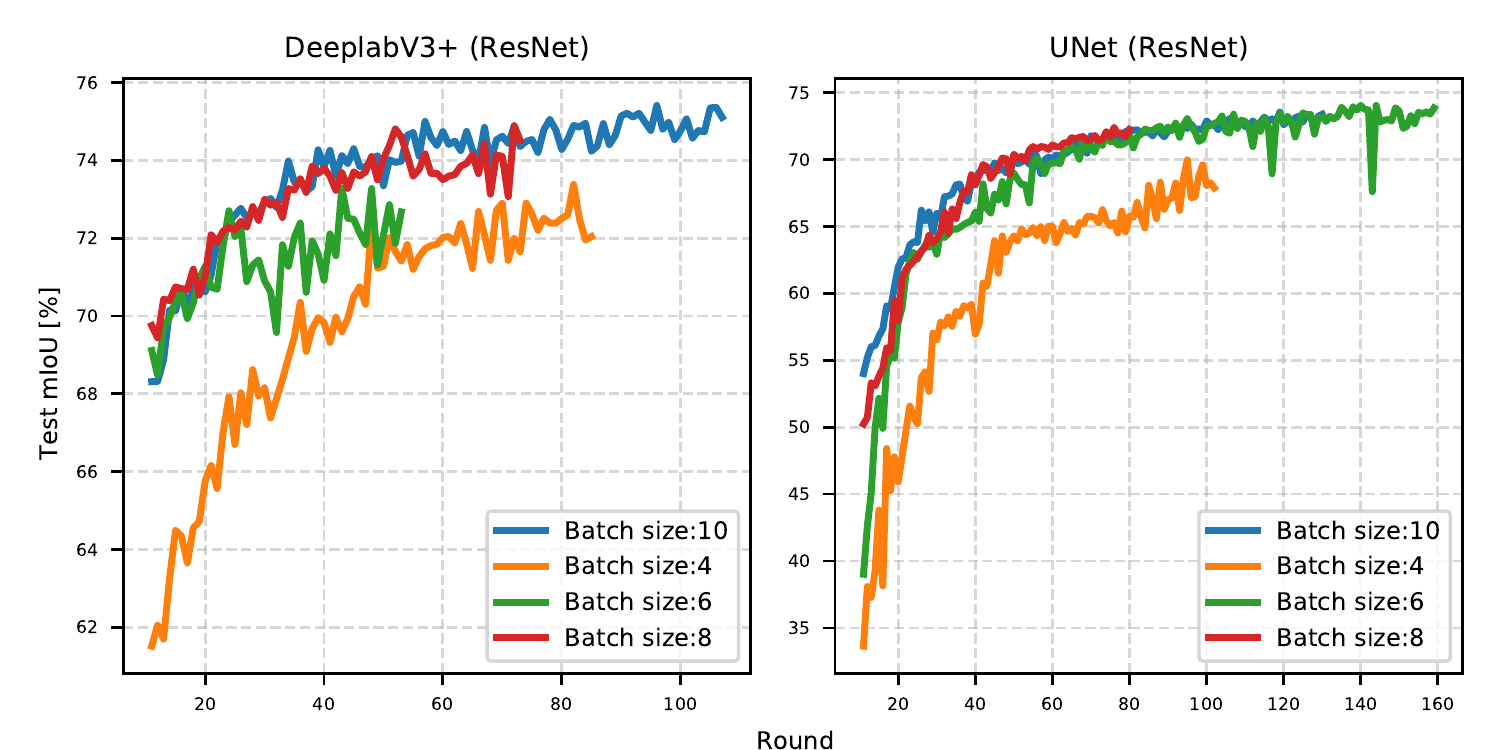}}\!\!\!\!\!\!
    \caption{Performance evaluation of segmentation tasks on Pascal VOC dataset. Figure (a): Comparing performance of DeeplabV3+ and UNet models with fine-tuning (FT) Resnet101 backbones against training from scratch (SFL)                     . Figure (b): Evaluating performance of DeeplabV3+ and UNet models on various batch sizes.}
    \label{fig:segment2}
\vspace{-0.5cm}
\end{figure*}

For the image segmentation, we train DeeplabV3+ and U-Net within the FedCV platform, in which the number of clients involved in each round of image segmentation are either 4 or 8. These studies are carried out on a computing cluster with a Quadro RTX 5000 graphics card. Each client has one GPU, with a 15.754 GB/s communication bandwidth.

Table \ref{tab:hyperparam_FedSeg} comprises a list of models and hyper-parameters we explored for evaluating performance of segmentation tasks in the federated setting. Note that we use following abbreviation throughout our analysis: \textbf{TT:} Training Type for Backbone. There are three strategies that we use for training the backbone. (i) Fine-Tuning \textbf{(FT)}:  We start with a ImageNet-pretrained backbone and fine-tune it for our task. (ii) Freezed Backbone \textbf{(FZ)}: Similarly to FT, we start with ImageNet\cite{russakovsky2015imagenet} pretrained backbone but do not train or fine-tune the backbone at all to save on computational complexity. (iii) Scratch Federated Learning \textbf{(SFL)}: Training the entire architecture end-to-end starting from scratch.

\begin{table}[ht]
\small
\centering
\begin{tabular}{|l|l|}
\hline
\textbf{Dataset} & Augemented PASCAL VOC\\ \hline
\textbf{Model} & DeeplabV3+, UNet\\ \hline
\textbf{Backbone} & ResNet-101, MobileNetV2 \\ \hline
\textbf{Backbone TT} & FT, FZ, SFL \\ \hline
\textbf{Batch Size Range} & 4 to 16 \\ \hline
\textbf{LR Range} & 0.0007 to 0.1 \\ \hline
\end{tabular}
\vspace{2 mm}
\caption{Dataset, models and hyper-parametesr choices for federated image segmentation task.}
\label{tab:hyperparam_FedSeg}
\end{table}

\begin{table}[ht]
\small
\centering
\begin{tabular}{|l|l|l|l|l|l|l|}
\hline
\textbf{Model} & \textbf{Backbone (TT)} & \textbf{DD} & \textbf{C} & \textbf{mIOU} \\ \hline
DeeplabV3+ & ResNet-101 (FT) & IID & 4 & 77.9\%  \\ \hline
DeeplabV3+ & ResNet-101 (FT) & N-IID & 4 & 76.47\%  \\ \hline
DeeplabV3+ & ResNet-101 (FT) & N-IID & 8 & 75.69\%  \\ \hline
DeeplabV3+ & ResNet-101 (SFL) & N-IID & 4 & 75.44\%  \\ \hline
DeeplabV3+ & ResNet-101 (FZ) & N-IID & 4 & 68.24\%  \\ \hline
DeeplabV3+ & MobileNetV2 (FT) & N-IID & 4 & 69.31\%  \\ \hline
UNet & ResNet-101 (FT) & IID & 4 & 75.14\%  \\ \hline
UNet & ResNet-101 (FT) & N-IID & 4 & 74.34\%  \\ \hline
UNet & ResNet-101 (FT) & N-IID & 8 & 73.65\%  \\ \hline
UNet & ResNet-101 (SFL) & N-IID & 4 & 74.2\%  \\ \hline
UNet & ResNet-101 (FZ) & N-IID & 4 & 51.19\%  \\ \hline
UNet & MobileNetV2 (FT) & N-IID & 4 & 66.14\%  \\ \hline
\end{tabular}
\vspace{2 mm}
\caption{Summary of test results on Pascal VOC dataset for federated image segmentation task. \textbf{DD:} Data Distribution Type. \textbf{N-IID}: Heterogeneous distribution with partition factor $\alpha$=0.5 \textbf{IID:} Homogeneous distribution. \textbf{C:} Number of Clients}
\label{tab:results}
\vspace{-0.3cm}
\end{table}

\subsubsection{Experimental Results}

In this section, we analyze and discuss our results for image segmentation tasks in the federated setting. We summarize our top results in Table \ref{tab:results} for a variety of training setups.

\noindent
\textbf{Backbone Training vs. Fine-Tuning.}\  Figure \ref{fig:segment2}(a) shows that pre-trained backbones coupled with fine-tuning results in only a slightly better performance (less than 2\%) compared to training from scratch, which indicates that while pre-trained backbones aid in federated image segmentation accuracy, they are not necessary. 
This finding opens the door to advanced tasks such as medical imaging, where pre-trained backbones may not be useful and end-to-end training from scratch is the only viable alternative.

\noindent
\textbf{Batch Size vs. Memory Trade-Off.}\  Figure \ref{fig:segment2}(b) and Table \ref{tab:Batch_Size} show that a smaller batch size, such as 4 instead of 10, reduces memory by roughly a factor of two while sacrificing nearly 2\% accuracy. This is an important trade off to make because edge devices in a federated learning setup may have constrained memory.


\begin{table}[ht]
\small
\centering
\begin{tabular}{|l|l|l|l|l|}
\hline
\textbf{Model} & \textbf{Backbone} & \textbf{BS} & \textbf{Memory} & \textbf{mIOU} \\ \hline
DeeplabV3+ & ResNet-101 & 4  & 6119M & 72.38\%  \\ \hline
DeeplabV3+ & ResNet-101 & 6  & 8009M & 73.28\%  \\ \hline
DeeplabV3+ & ResNet-101 & 8  & 10545M & 74.89\%  \\ \hline
DeeplabV3+ & ResNet-101 & 10  & 13084M & 75.5\%  \\ \hline
UNet & ResNet-101 & 4  & 6032M & 71.54\%  \\ \hline
UNet & ResNet-101 & 6  & 8456M & 71.89\%  \\ \hline
UNet & ResNet-101 & 8  & 10056M & 72.4\%  \\ \hline
UNet & ResNet-101 & 10  & 12219M & 73.55\%  \\ \hline
\end{tabular}
\vspace{2 mm}
\caption{Performance and memory analysis for various batch size of segmentation models on Pascal VOC Dataset. \textbf{BS:} Batch Size}
\label{tab:Batch_Size}
\vspace{-0.3cm}
\end{table}



\begin{figure*}[htb!]
\small
   \centering
   \!\!\!\!\!\!\subfigure[Experiments on various partition factors]{\includegraphics[width=9cm]{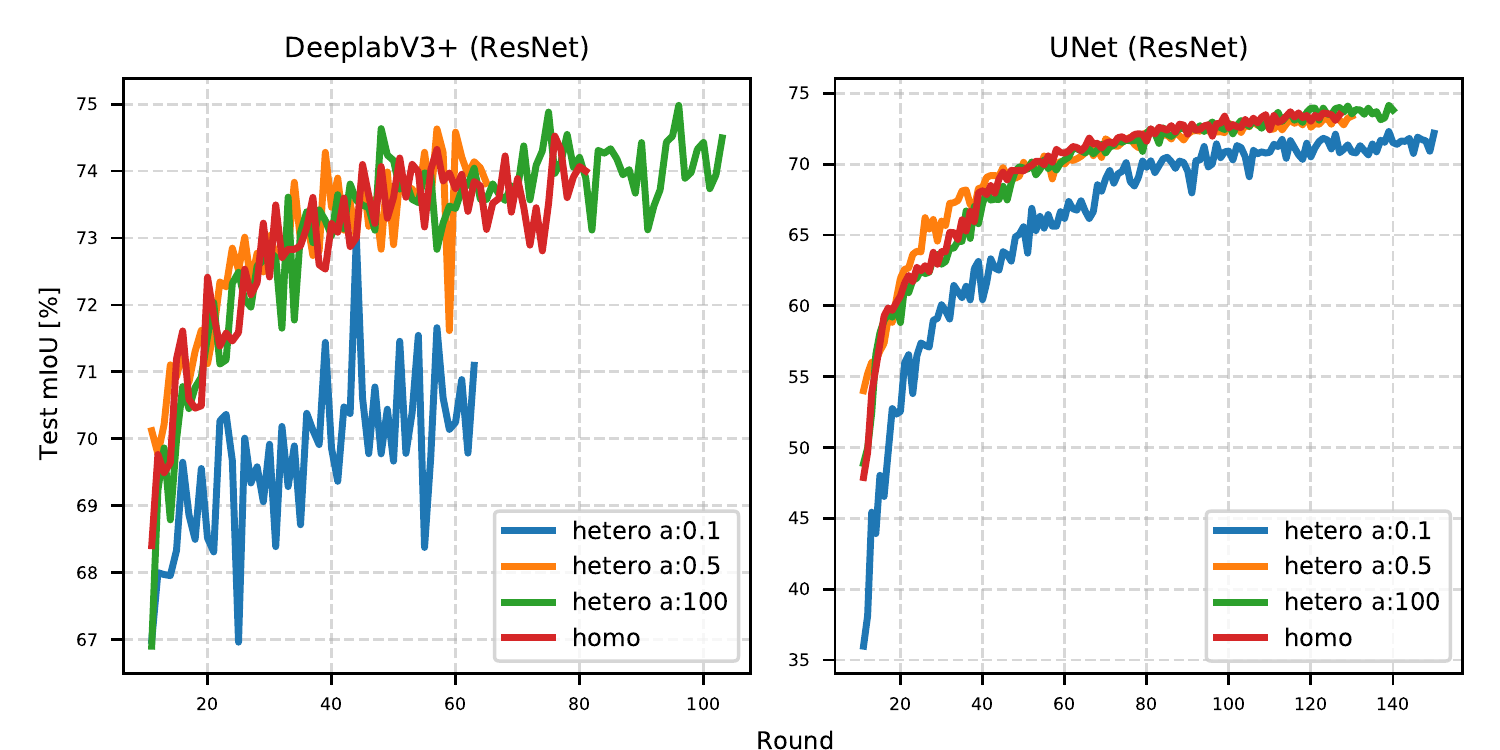}}\!\!  
   \!\!\subfigure[Experiments on varying number of clients ]{\includegraphics[width=9cm]{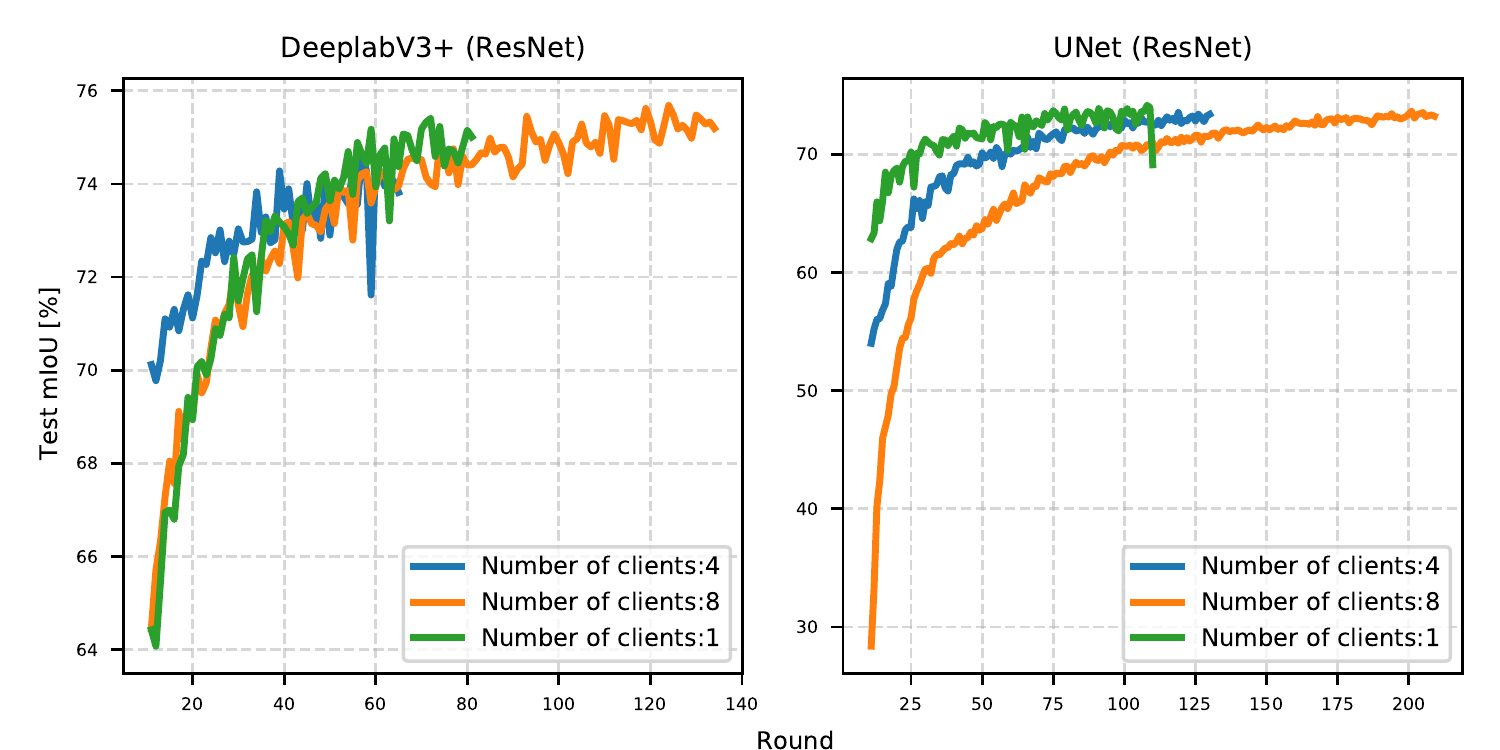}}\!\!\!\!\!\!
    \caption{Performance evaluation of segmentataion task on Pascal VOC dataset. Figure (a): Evaluating performance of DeeplabV3+ and UNet models with Resnet101 as a backbone on various partition factors (a).Figure (b): Evaluation performance of DeeplabV3+ and UNet models with Resnet-101 as backbone on varying number of clients.}
    \label{fig:data_client_partition}
\end{figure*}

\begin{table*}[h!]
\small
\centering
\begin{tabular}{|l|l|l|l|l|l|l|}
\hline
\textbf{Model} & \textbf{Backbone (TT)} & \textbf{Dataset} & \textbf{Params} & \textbf{FLOPS} & \textbf{Memory (BS)} & \textbf{Total Time} \\ \hline
DeeplabV3+ & ResNet-101 (FT) & PASCAL VOC & 59.34M & 88.85G & 13084M (10) & 14.16h \\ \hline
DeeplabV3+ & ResNet-101 (FZ) & PASCAL VOC & 16.84M & 88.85G & 7541M (16) & 23.59h \\ \hline
DeeplabV3+ & MobileNetV2 (FT) & PASCAL VOC & 5.81M & 26.56G & 12104M (16) & 20.5h \\ \hline
UNet & ResNet-101 (FT) & PASCAL VOC & 51.51M & 62.22G & 12219M (10) & 14.5h \\ \hline
UNet & ResNet-101 (FZ) & PASCAL VOC & 9.01M & 62.22G & 7687M (16) & 51.11h \\ \hline
UNet & MobileNetV2 (FT) & PASCAL VOC & 7.91M & 14.24G & 11706M (16) & 22.03h \\ \hline
\end{tabular}
\vspace{2 mm}
\caption{System performance chart of segmentation network architectures we considered. \textbf{TT:} Training Type. \textbf{BS:} Batch Size}
\label{tab:SystemAnalysis}
\vspace{-0.3cm}
\end{table*}

\noindent
\textbf{Data distribution impact analysis.}\ For various partition values $\alpha$, Figure \ref{fig:heatmap} depicts the distribution of classes among clients. Even when the partition factor changes from totally homogeneous to extremely heterogeneous, as shown in Figure \ref{fig:data_client_partition} (a), the accuracy only degrades by about 2\%. This further demonstrates that federated segmentation learning can instill enough generalization capability in local clients to allow them to perform well on unknown data, obviating the need for centralized or widely distributed data.




\noindent
\textbf{Resiliency in the face of increasing clients}\  The number of rounds needed for the model to converge increases as the number of clients increases (see figure \ref{fig:data_client_partition}(b)). When compared to smaller client sizes, which are theoretically expected to perform better since each local client has more data points to train on, it has little effect on final accuracy after a sufficient number of rounds.

\subsubsection{System Performance Analysis}

\begin{figure}[htp]
\small
    \centering
    \includegraphics[width=7cm]{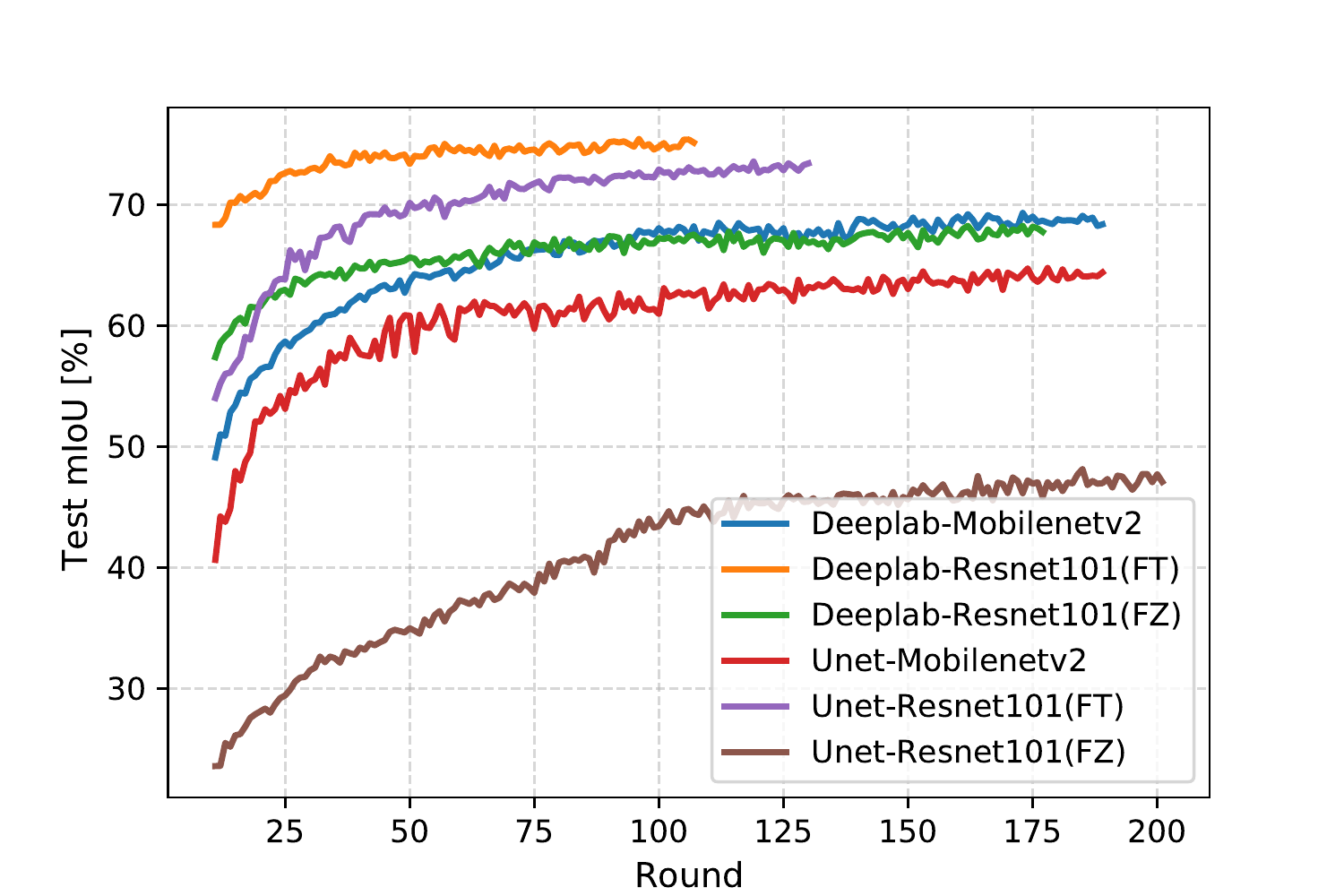}
    \caption{Performance comparison of DeeplabV3+ and UNet with ResNet101 and MobileNetV2 as backbones. DeeplabV3+ (Resnet101) reaches a better accuracy compared to other alternatives. \textbf{FT:} Fine-Tuning Backbone. \textbf{FZ:} Freezed Backbone}
    \label{fig:model_backbone}
\end{figure}
ResNet is one of the most widely used backbones for encoder-decoder architecture in image segmentation tasks; however, it has a high computing cost that many edge devices might not be able to bear. There are two obvious ways to trim the cost down: (i) Freezing the pre-trained backbone; (ii) Plugging computationally efficient backbone (Eg. MobileNetV2).
Figure \ref{fig:model_backbone}  depicts the performance variance when one of the two described strategies is applied for backbones in DeeplabV3+ and UNet architectures for federated image segmentation. When compared to every other mix, the accuracy of ResNet-101 backbone is demonstrably higher. On the other hand, as shown in Table \ref{tab:SystemAnalysis}, the alternatives are extremely efficient at the cost of performance degradation.





\subsection{Object Detection}
\subsubsection{Implementation Details}
For object detection, we use pre-trained YOLOV5 for federated learning experiments with the FedAvg algorithm. The client number we used include 4 and 8 for performance comparison. Each client was run at one GPU (NVIDIA V100). The metric in our experiments is mAP@0.5 (mean average precision with a threshold of 0.5 for IOU).

\begin{figure}[htb!]
\small
   \centering
   \!\!\!\!\subfigure[Different learning rates]{\includegraphics[height=3.5cm, width=4.5cm]{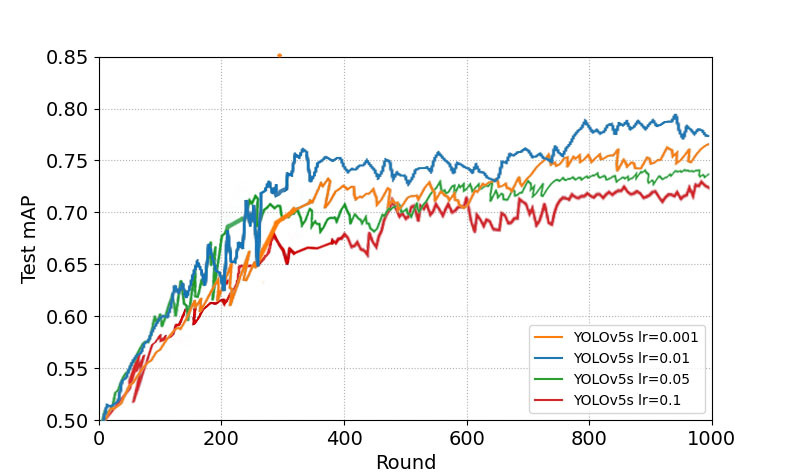}}\!\!\!\!\!  
   \!\!\!\!\!\subfigure[Different number of clients]{\includegraphics[height=3.5cm, width=4.5cm]{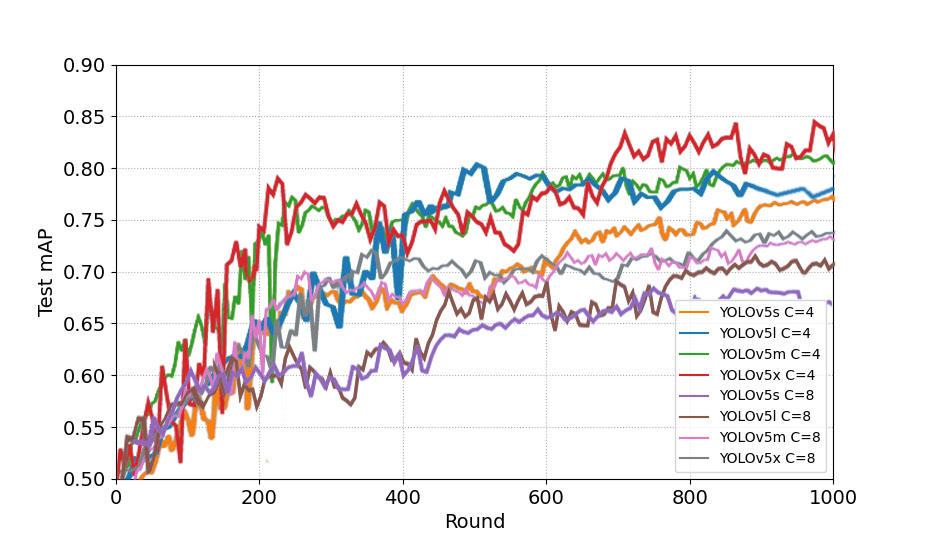}}\!\!\!\!\!\!
    \caption{Experiments on detection tasks on varying learning rates and number of clients. Figure (a): Non-IID data comparsion with different learning rate. Figure (b): Non-IID data comparsion with different number of clients. }
    \label{fig:fedobd_1_2}
\end{figure}

\subsubsection{Experimental Results}

\noindent
\textbf{Learning rate.} In the federated setting, different learning rates are evaluated. While keeping the other hyper-parameters (e.g., client number is set to 4), we notice that \textit{lr}=0.01 can have a better result compared to the other choices from Figure \ref{fig:fedobd_1_2} (a).

\noindent
\textbf{Non-I.I.D. evaluation.} For the client numbers 4 and 8, we use the partition method introduced in the appendix to obtain synthetic federated datasets. We found that when using YOLOv5, it is difficult for FedAvg to reach the same results as that of centralized for Non-IID dataset. Figure \ref{fig:fedobd_1_2} shows there is a large gap between centralized training and FedAvg-based federated training. 
In centralized training of YOLOv5, test mAP of all four model variants is over 0.95 \cite{YOLOv5}, whereas the best accuracy in the federated setting is smaller than 0.85. The main reason is that the optimizer and training tricks used in centralized training could not be directly transplanted to the FL framework, indicating that further research for object detection in the federated setting is required.

\noindent
\textbf{Evaluation on different number of clients.}  We also show the performance with different clients among 4 models in Figure \ref{fig:fedobd_1_2} (b). Results showing that \textit{C} = 8 has a lower performance compared to the \textit{C} = 4. 

\noindent
\textbf{System performance analysis.} Table \ref{tab:hyperparam_FedSOJ} summarizes the system performance of four different model variants. We can see that as the network structure depth and width increased among the four models, the model performed well with a better mAP.

\begin{table}[h!]
\small
\centering
\begin{tabular}{|l|l|l|l|l|l|}
\hline
\textbf{Model} &  \!\!\textbf{Layers}\!\! &\!\! \textbf{Parameters} \!\! & \!\!\textbf{FLOPS} \!\!&\!\! \textbf{Total Time}\!\!\\ \hline
\!\!\!YOLOv5s \!\! & 283 & 7.27M  & 17.1G & 25.1h\\ \hline
\!\!\!YOLOv5m \!\!& 391 & 21.4M  & 51.4G & 49.3h \\ \hline
\!\!\!YOLOv5l \!\!& 499 & 47.1M  & 115.6G & 73.5h \\ \hline
\!\!\!YOLOv5x \!\!& 607 & 87.8M   & 219.0G & 92.4h \\ \hline
\end{tabular}
\vspace{2 mm}
\caption{System performance of YOLOv5}
\label{tab:hyperparam_FedSOJ}
\end{table}

\section{Conclusion}\label{conclusion}
In this work, we propose an easy-to-use federated learning framework for diverse computer vision tasks, including image classification, image segmentation, and object detection, dubbed \texttt{FedCV}. We provide several non-IID benchmarking datasets, models, and various reference FL algorithms. We hope that \texttt{FedCV} can open doors for researchers to develop new federated algorithms for various computer vision tasks. 

FedML Ecosystem \cite{He2020FedMLAR} aims to provide a one-stop scientific research platform through FedML Ecosystem and finally realize trustworthy ML/AI, which is more secure, scalable, efficient, and ubiquitous. FedCV serves as one of the key components of FedML ecosystem. The other important applications include FedNLP \cite{lin2021fednlp}, FedGraphNN \cite{he2021fedgraphnn}, and FedIoT \cite{zhang2021federated}.

{\small
\bibliographystyle{ieee_fullname}
\bibliography{egbib}
}

\clearpage
\section*{Appendix}

In the appendix, we provide more details of the benchmark suite and experiments.

\subsection{Benchmark Suite}

\subsubsection{Dataset}

\textbf{CIFAR-100.} The CIFAR-100 dataset \cite{cifar100} has 100 classes, of which each contains 600 images with size 32 $\times$ 32. 

\textbf{Google Landmarks Dataset 23k (GLD-23K)} is a subset of Google Landmark Dataset 160k \cite{gld}. This GLD-23K dataset includes 203 classes, 233 clients, and 23080 images. We follow the setting of GLD-23K from Tensorflow federated \cite{gldtf}. We also provide the data loader for GLD-160K  in our source code.

\textbf{PASCAL VOC - Augmented.} We use the augmented PASCAL VOC dataset with annotations from 11355 images \cite{BharathICCV2011}. These images are taken from the original PASCAL VOC 2011 dataset, which contains 20 foreground object classes and one background class. 

\textbf{COCO \cite{lin2014microsoft}} is a dataset for detecting and segmenting objects found in everyday life through extensive use of Amazon Mechanical Turk.

\subsubsection{Non-I.I.D. Partition and Distribution Visualization}
For GLD-23K, we follow the setting of  GLD-23K from Tensorflow federated \cite{gldtf}, which means that the number of total clients is 233 for GLD-23K.

We make use of Latent Dirichlet Allocation (LDA) \cite{LDA} method to partition CIFAR-100 and PASCAL VOC into non-I.I.D. dataset. The settings of our non-I.I.D. partition can be referred to table \ref{tab:summary_dataset_partition}.

\begin{figure}[htp]
\small
    \centering
    \includegraphics[height=14cm,width=8.5cm]{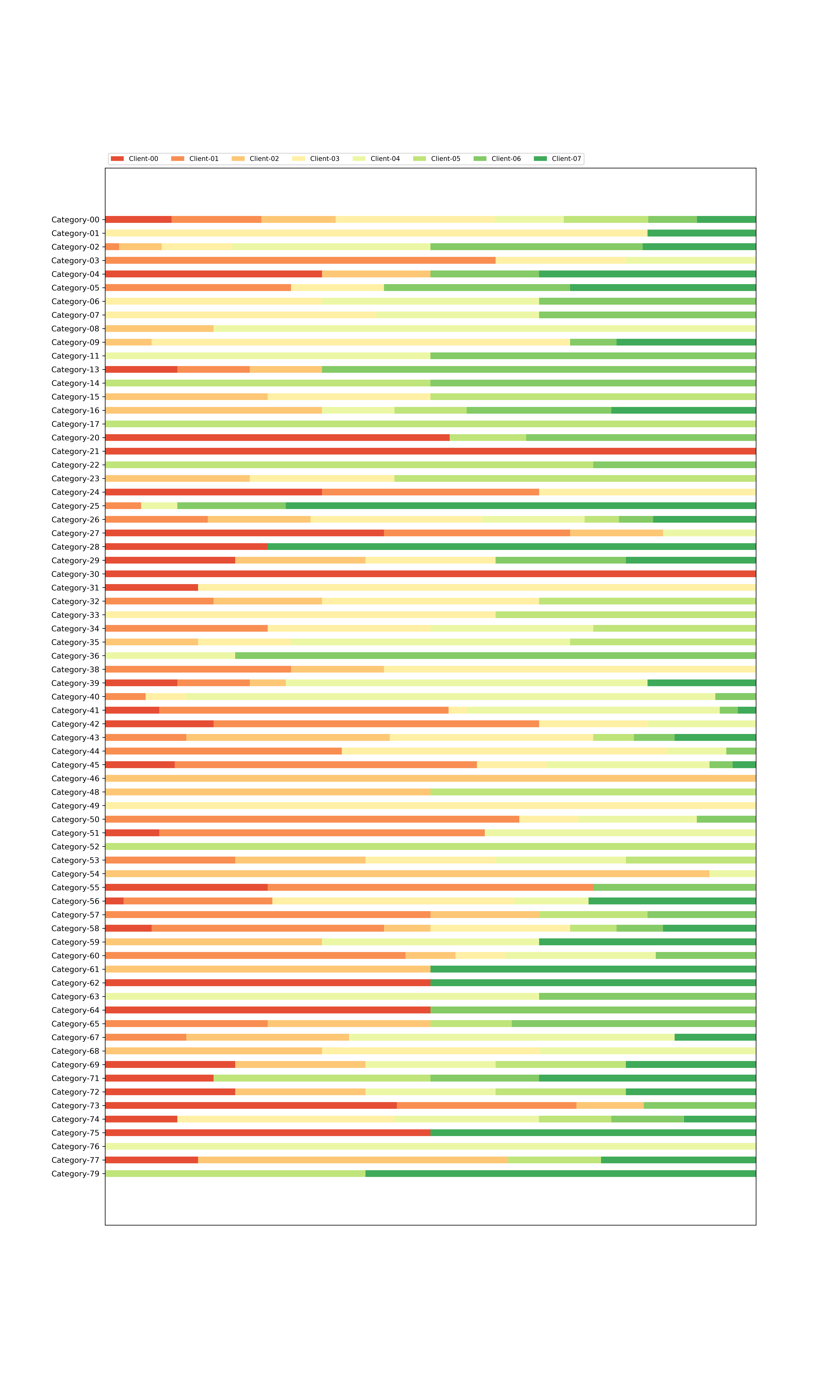}
    \caption{Non-IID data distribution on COCO dataset. Here we show the figure that COCO data has been distributed to 10 clients in a non-partition way. For the COCO dataset non-IID distribution, we have 80 categories and they are assigned to different clients showed by different color. }
    \label{fig:coco_noniid}
\end{figure}

Comparing to the classification and segmentation task with one picture has one label using the LDA partition method, the object detection task always has several labels on one image. In this case, we take a different partition method. First, we calculate the frequency of each object in all images (one object is only counted one time even it occurs more than one time on one image). Second, we sort the objects' frequency and put the object with the highest frequency into one client. Finally, we sort the remaining pictures and repeat the first step until all the images have been assigned to the clients. Figure \ref{fig:coco_noniid} shows the COCO non-IID data distribution on 8 clients. Different color represents different clients, and we could see every categories label data are non-IID distributed to different clients. 

\begin{figure*}[htb!]
\small
\vspace{-0.2cm}
   \centering
   \!\!\!\!\!\!\!\!\subfigure[CIFAR-100 with $\alpha=0.1$]{\includegraphics[width=0.255\textwidth]{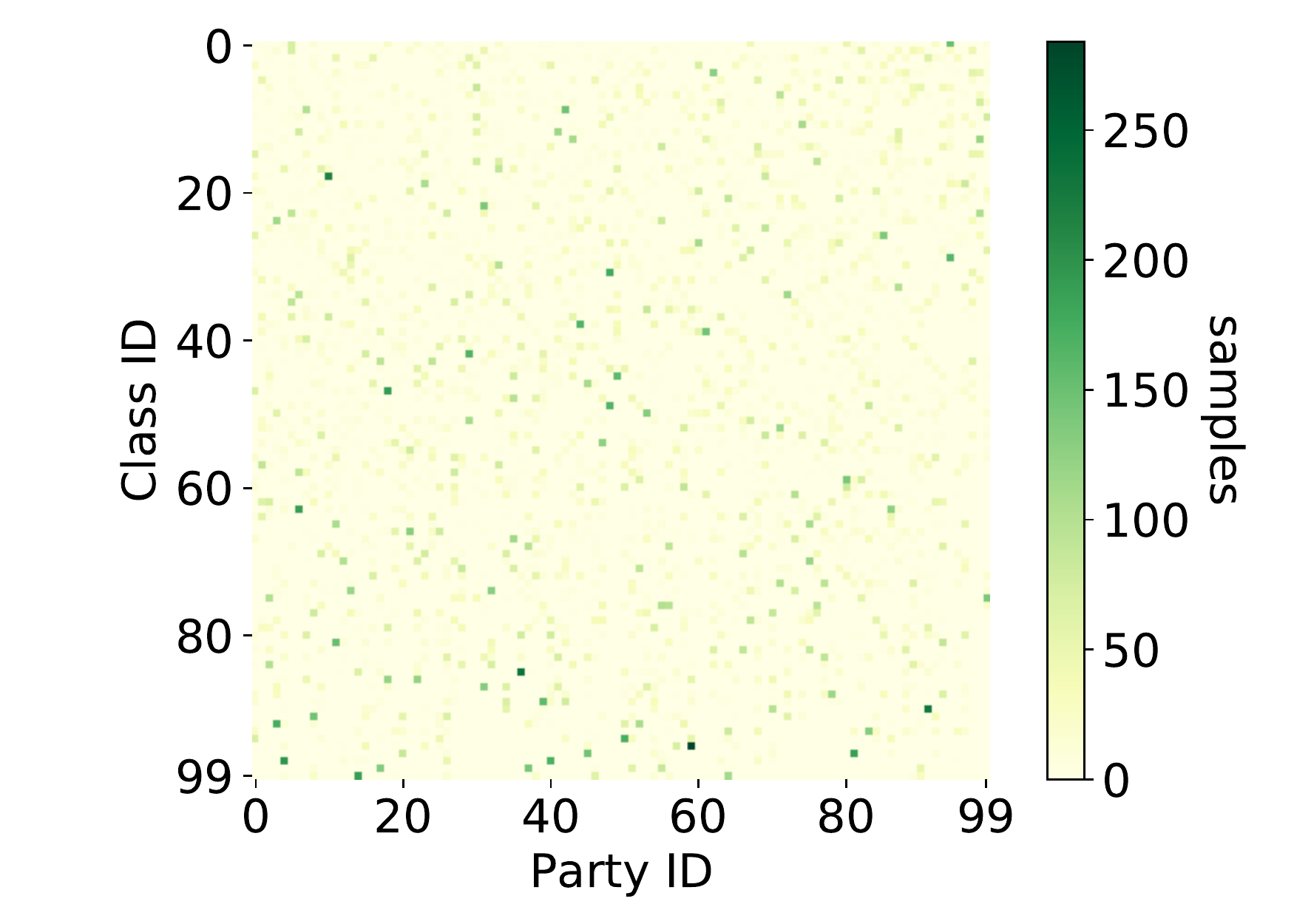}}\!\!\!\!
   \subfigure[CIFAR-100 with $\alpha=0.5$]{\includegraphics[width=0.255\textwidth]{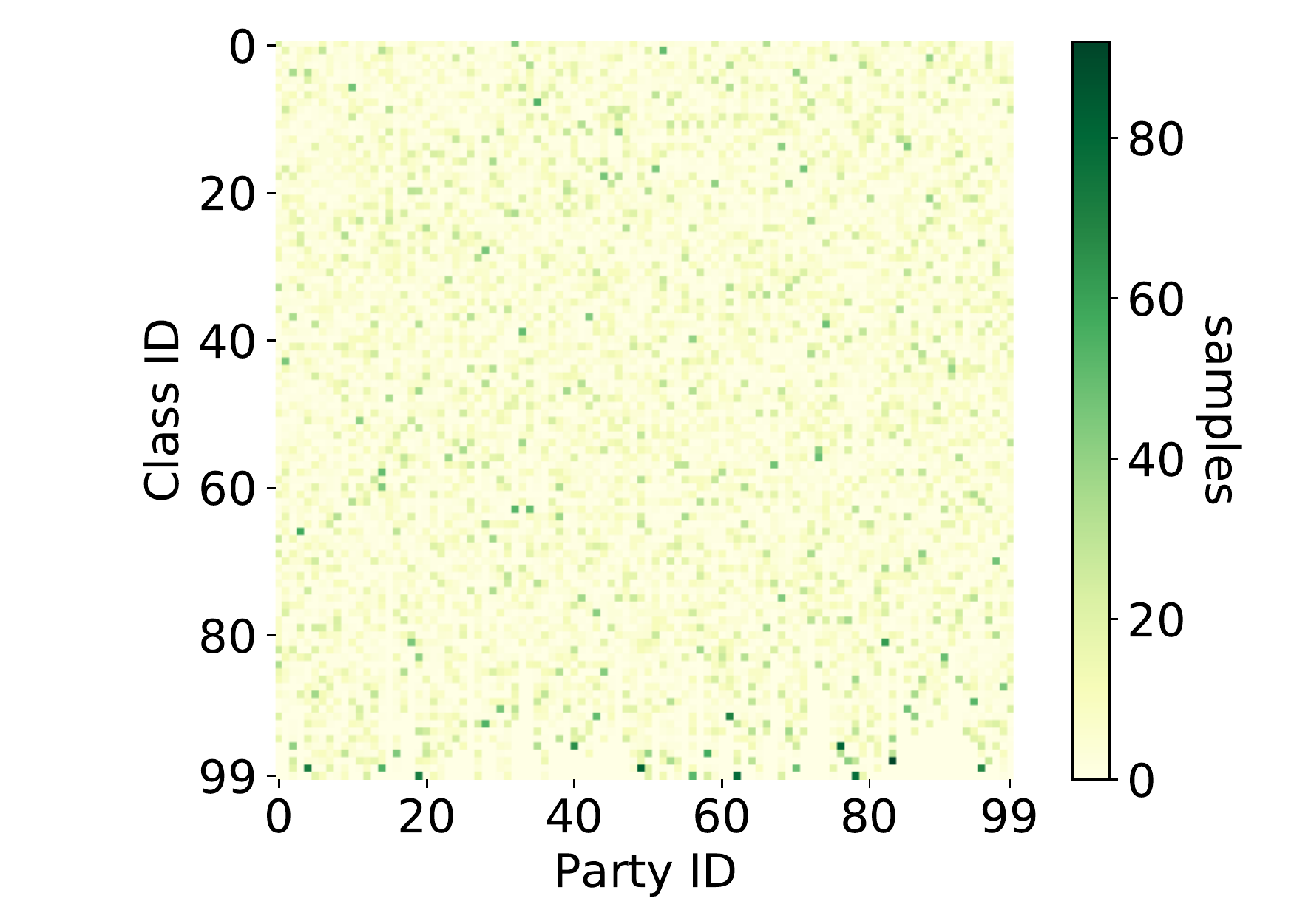}}\!\!\!\!
   \subfigure[CIFAR-100 with $\alpha=100.0$]{\includegraphics[width=0.255\textwidth]{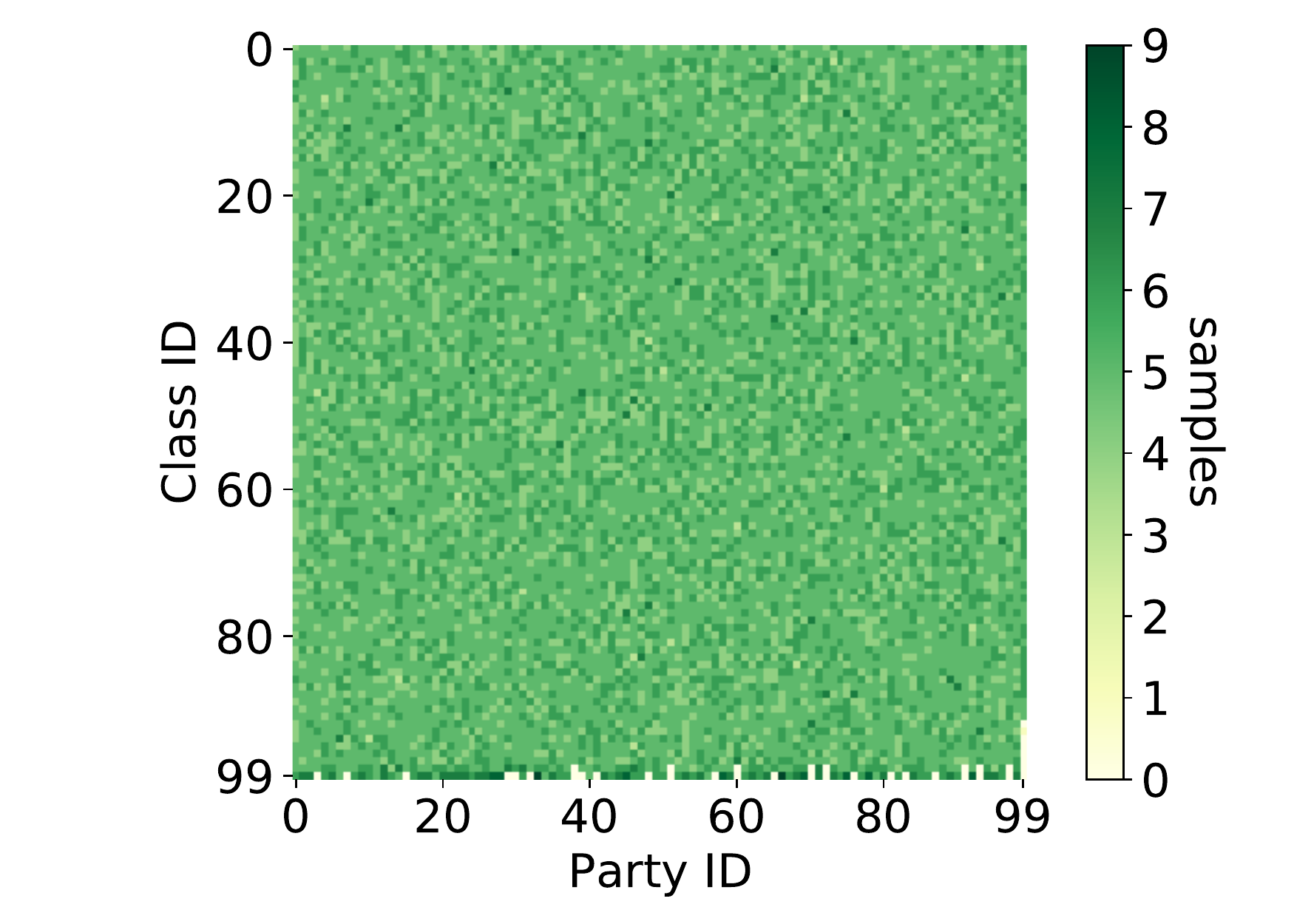}}\!\!\!\!
   \subfigure[GLD-23K]{\includegraphics[width=0.255\textwidth]{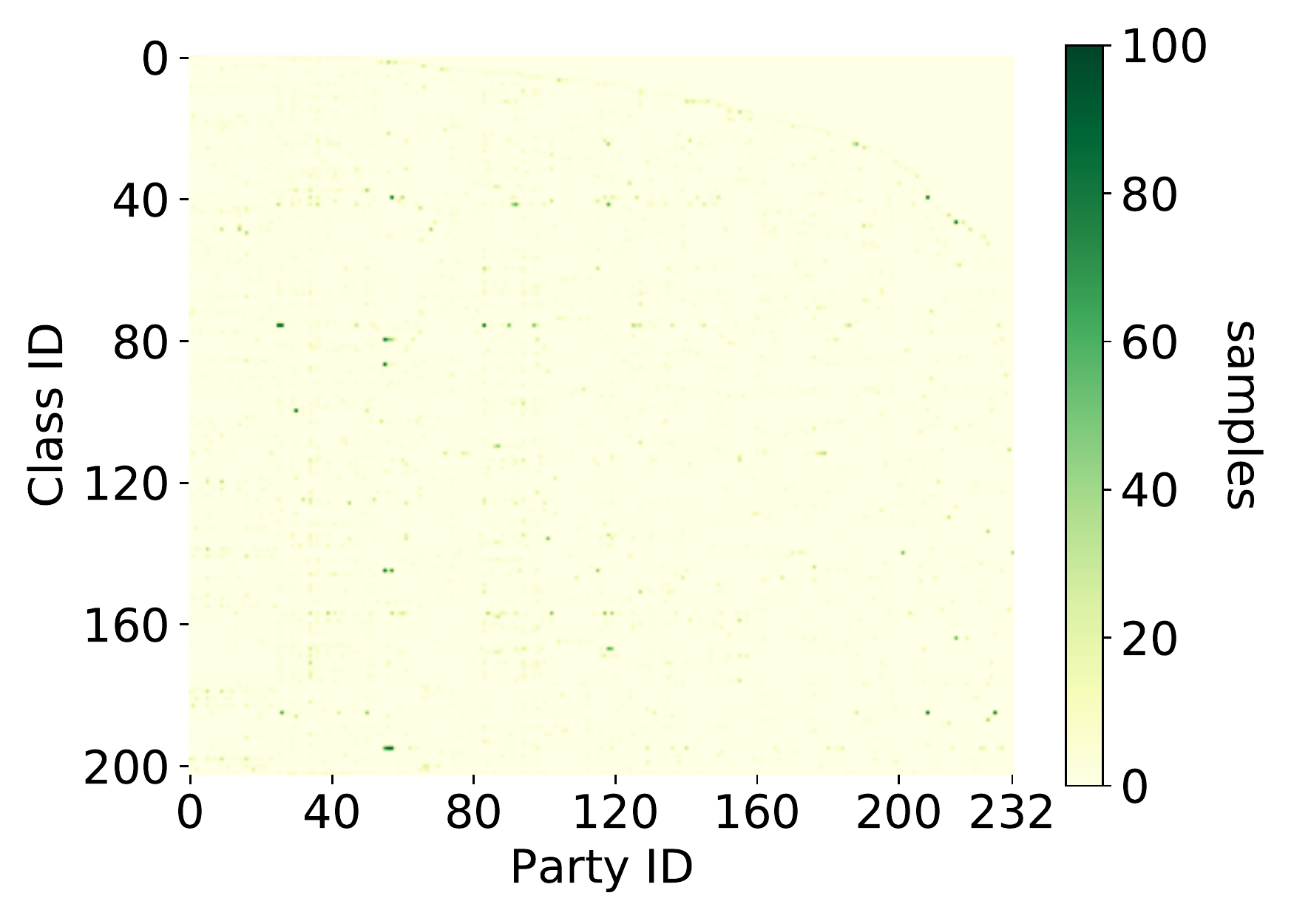}}\!\!\!\!\!
    \caption{Visualization on classification tasks. 
    Figure (a): Visualization of CIFAR-100 distribution on 100 clients with $\alpha=0.1$.
    Figure (b): Visualization of CIFAR-100 distribution on 100 clients with $\alpha=0.5$.
    Figure (c): Visualization of CIFAR-100 distribution on 100 clients with $\alpha=100.0$.
    Figure (d): Visualization of GLD-23K distribution on 233 clients.}
    \label{fig:Visualization_classification}
\end{figure*}

Figure \ref{fig:Visualization_classification}(a)-(c) show the visualization of non-IID CIFAR-100 with different $\alpha$. When the $\alpha$ increases, the similarity of data distribution becomes higher. Figure \ref{fig:Visualization_classification}(d) visualize the data distribution of GLD-23K on 233 clients. For non-IID CIFAR-100 with a low $\alpha$ value and GLD-23K, we can see that the visualization matrix is sparse. Some clients have many samples of some labels, but few samples of some other labels. This makes training become much harder.

\subsubsection{Models}

\textbf{EfficientNet} \cite{efficientnet} and \textbf{MobileNet-V3}  \cite{mobilenet} are two light weighted convolutional neural networks. They achieves the goal of improving accuracy and greatly reducing the amount of model parameters and calculations. In this paper, we use EfficientNet-b0 and MobileNet-V3 Large to conduct experiments.

\textbf{Vision Transformer (ViT) \cite{ViT}}  is a novel neural network exploiting transformer into Computer Vision and attain excellent results compared to state-of-the-art convolutional networks. We use ViT-B/16 to conduct experiments.

\textbf{MobileNetV2}is a lightweight Convolutional Neural Network. It is primarily designed to support running neural networks in mobile and edge devices that have severe memory constraints. It implements Inverted Residuals concept where the residual connections are used between bottleneck layers. It also applies Linear Bottlenecks and Depthwise Separable Convolutions concept. In our implementation, we use a network that was pre-trained on the ImageNet dataset.

\begin{figure}[!htp]
    \centering
    \includegraphics[width=8.5cm]{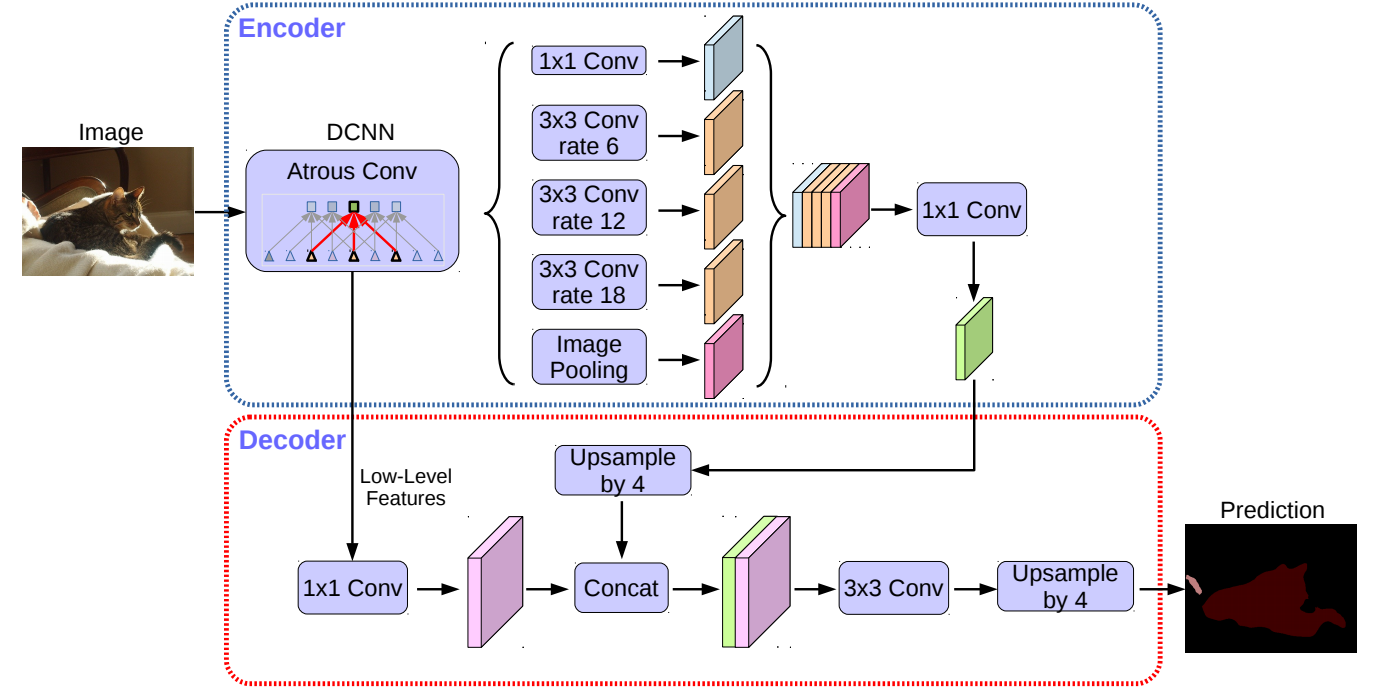}
    \caption{DeeplabV3+ Architecture taken from \cite{DeepLabV3+}}
    \label{fig:DeepLabV3Plus}
\end{figure}

\textbf{DeepLabV3+} is a neural network that employs two main principles Atrous Convolutions, Depthwise Separable Convolutions, and Encoder-Decoder architecture (as shown in Figure \ref{fig:DeepLabV3Plus} which attain great performance on Image Segmentation. In our experiments, we exploit MobileNet-V2 and ResNet-101 as two kinds of backbones of DeepLabV3+ to conduct our experiments.

\begin{figure}[!htp]
    \centering
    \includegraphics[width=8.5cm]{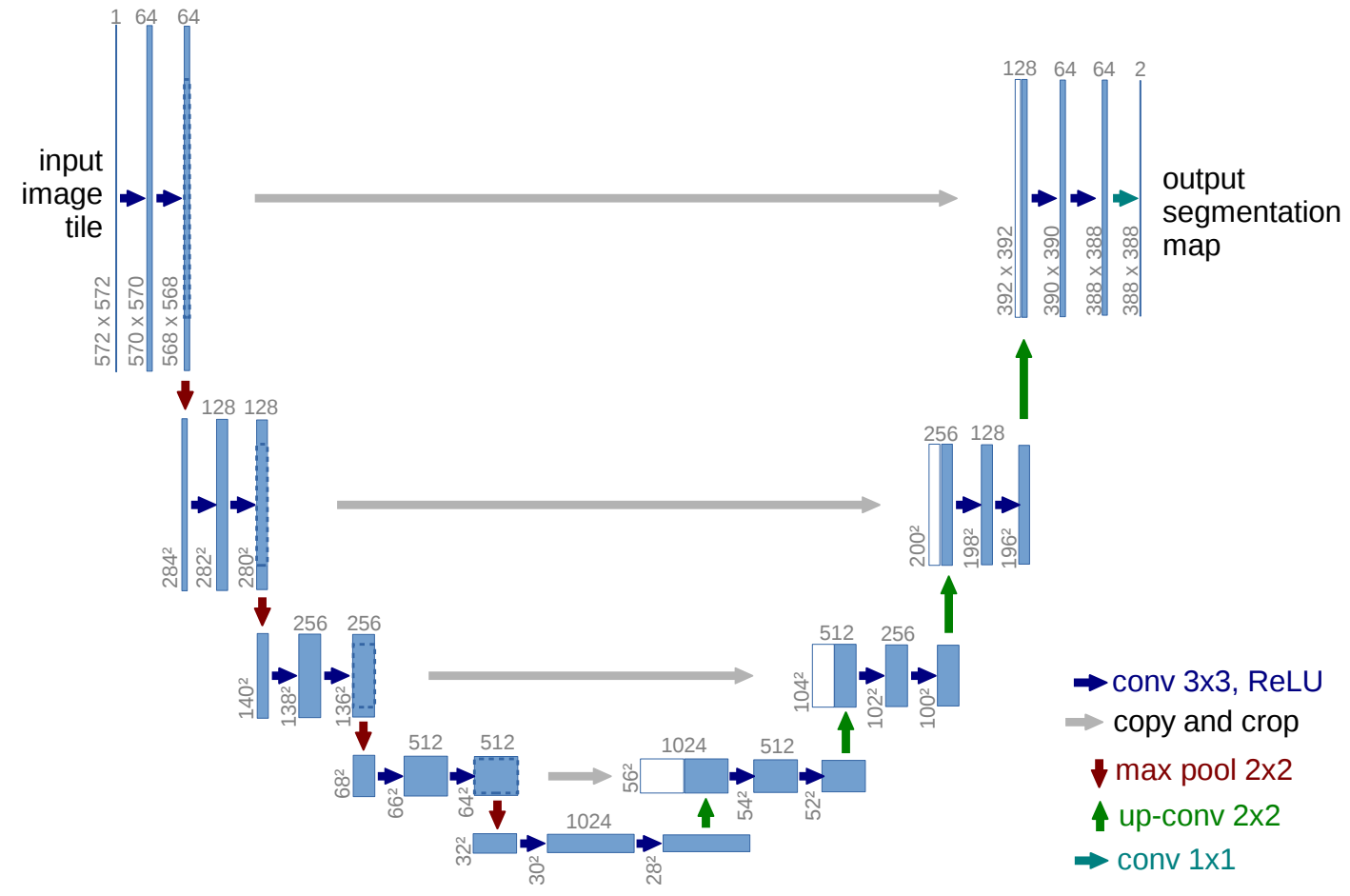}
    \caption{U-Net Architecture taken from \cite{U-Net}}
    \label{fig:unet}
\end{figure}

\textbf{U-Net \cite{U-Net}} is a Convolutional Neural Network that follows an Encoder-Decoder architecture pattern. As shown in Figure \ref{fig:unet}, U-Net does not need a backbone network to perform segmentation. However, we have experimented with Resnet-101 and MobileNetV2 pre-trained backbones during experimentation to improve the segmentation output further.

\textbf{YOLOv5 \cite{YOLOv5}} is an optimized version of YOLOv4 \cite{bochkovskiy2020yolov4}. It outperforms all the previous versions and gets near to EfficientDet Average Precision(AP) with higher frames per second (FPS). Four network models (Yolov5s, Yolov5m, Yolov5l, Yolov5x) with different network depths and widths cater to various applications. Here we use these four models as the pre-trained models.

\subsubsection{FL Algorithms}

\begin{table*}[!htp]
\small
\centering
\begin{tabular}{|l|r|l|r|} 
\hline
Dataset              & \multicolumn{1}{l|}{clients} & Partition                                       & \multicolumn{1}{l|}{Num of labels}  \\ 
\hline
CIFAR-100            & 100                          & LDA with $\alpha$ in $\left \{0.1, 0.5, 100 \right\}$ & 100                                 \\ 
\hline
GLD-23K              & 233                          & Default                                         & 203                                 \\ 
\hline
augmented PASCAL VOC & 4, 8                   & LDA with $\alpha$in $\left \{0.1, 0.5, 100 \right\}$ & 21                                  \\ 
\hline
COCO                 & 8                            & Our partition                                  & 80                                  \\
\hline
\end{tabular}
\caption{Summary of Dataset partition.}
\label{tab:summary_dataset_partition}
\end{table*}

\textbf{Optimizer.} Unlike vanilla SGD in FedAvg \cite{brendan2016}, here we use Momentum SGD to optimize models. For centralized training, our Momentum SGD is the same as the traditional Momentum SGD. However, for FedAvg, the accumulated gradient will be cleared when clients receive the global model from the server. The momentum coefficient is 0.9 for all experiments.

\textbf{Learning Rate Scheduler.} For centralized training on all tasks, we exploit linear learning rate decay each epoch. For FedAvg, we do not use a learning rate scheduler on all tasks. However, we try it on image classification in order to evaluate its effect. Note that the learning rate decay is based on the communication round, not the local epochs as the traditional training process.

\textbf{FedAvg Training.} 
We use the FedAvg \cite{brendan2016} algorithm to conduct federated learning. We list our FedAvg training settings in table \ref{tab:fedavg_settings}, in which $C$ means the number of clients participating in calculation per round, and $E$ means the number of local epochs per round.

\begin{table}[!htp]
\small
\centering
\begin{tabular}{|l|r|r|} 
\hline
Task                 & C & E  \\ 
\hline
Image Classification & 10                     & 1                       \\ 
\hline
Image Segmentation   & 4, 8                   & 1, 2                    \\ 
\hline
Object Detection     & 4, 8                   & 1                       \\
\hline
\end{tabular}
\caption{Summary of FedAvg settings.}
\label{tab:fedavg_settings}
\end{table}

\textbf{Image Transform.} For federated learning, it can not be assumed that clients know data distribution of other clients. So we use the average RGB value as $\left[0,5, 0.5, 0.5 \right]$ and standard deviation as $\left[0,5, 0.5, 0.5 \right]$, instead of them of all images.

\textbf{Layer-wise Learning}
Both the models for image segmentation task have encoder-decoder architecture. For encoder layers, we employ ImageNet-1K pre-trained backbones, and hence they are to be just fine-tuned. The decoder layers are trained from scratch. To enforce this set-up, we modify our learning rates layer-wise so that the decoder layers get a learning rate 10 times more than the encoder layers.

\subsection{More Experimental Results and Hyper-parameters}
For image classification, we list all experiment results and the corresponding hyper-parameters in table \ref{tab:summary_of_cifar100_efficientnet}, \ref{tab:summary_of_cifar100_mobilenet}, \ref{tab:summary_of_gld23k_efficientnet}.


\begin{table}[!htp]
  \centering
  \fontsize{9}{9}\selectfont
  \begin{threeparttable}
    \begin{tabular}{cccccc}
    \toprule
Data                   & Opt.                   & Sched.                  & LR    & SFL. Acc              & FT. Acc  \\ 
\hline
\multirow{5}{*}{Cent.} & \multirow{5}{*}{M SGD} & \multirow{5}{*}{Linear} & 0.003 &  & 0.6011   \\ 
                       &                        &                         & 0.01  & & 0.6058   \\ 
                       &                        &                         & 0.03  &   & 0.5931   \\ 
                       &                        &                         & 0.1   &  & 0.6055   \\ 
                       &                        &                         & 0.3   &  & 0.215    \\ 
\hline
\multirow{7}{*}{a=0.1} & \multirow{5}{*}{M SGD} & \multirow{3}{*}{No}     & 0.003 & 0.2239                & 0.4295   \\ 
                       &                        &                         & 0.01  & 0.2306                & 0.2651   \\ 
                       &                        &                         & 0.03  & 0.03025               & 0.37     \\ 
\cline{3-6}
                       &                        & \multirow{2}{*}{Linear} & 0.01  & 0.2384                & 0.3104   \\ 
                       &                        &                         & 0.03  & 0.2469                & 0.2676   \\ 
\cline{2-6}
                       & \multirow{2}{*}{SGD}   & \multirow{2}{*}{Linear} & 0.01  & 0.1358                & 0.1451   \\ 
                       &                        &                         & 0.03  & 0.1779                & 0.3203   \\ 
\hline
\multirow{7}{*}{a=0.5} & \multirow{5}{*}{M SGD} & \multirow{3}{*}{No}     & 0.003 & 0.3315                & 0.5112   \\ 
                       &                        &                         & 0.01  & 0.4092                & 0.5502   \\ 
                       &                        &                         & 0.03  & 0.433                 & 0.4841   \\ 
\cline{3-6}
                       &                        & \multirow{2}{*}{Linear} & 0.01  & 0.3045                & 0.5329   \\ 
                       &                        &                         & 0.03  & 0.3997                & 0.53     \\ 
\cline{2-6}
                       & \multirow{2}{*}{SGD}   & \multirow{2}{*}{Linear} & 0.01  & 0.1847                & 0.4479   \\ 
                       &                        &                         & 0.03  & 0.2861                & 0.5366   \\ 
\hline
\multirow{3}{*}{a=100} & \multirow{3}{*}{M SGD} & \multirow{3}{*}{No}     & 0.003 & 0.3732                & 0.6158   \\ 
                       &                        &                         & 0.01  & 0.4176                & 0.5749   \\ 
                       &                        &                         & 0.03  & 0.4006                & 0.5527   \\
    \bottomrule
    \end{tabular}
    \end{threeparttable}
\caption{Summary of test accuracy on CIFAR-100 with EfficientNet-b0. In this table, Opt. represents optimizer, Sched. means learning rate scheduler, SFL. means \textbf{scratch federated learning}, i.e. not loading a pretrained model, and FT. means \textbf{fine-tining}, i.e. loading a pretrained model. In other tables, we also use these abbreviations.}
\label{tab:summary_of_cifar100_efficientnet}
\vspace{0.1cm}
\end{table}

\begin{table}[!htp]
  \centering
  \fontsize{9}{9}\selectfont
  \begin{threeparttable}
    \begin{tabular}{cccccc}
    \toprule
Data                   & Opt.                   & Sched.                  & LR &  SFL. Acc &  FT. Acc  \\ 
\hline
\multirow{5}{*}{Cent.} & \multirow{5}{*}{M SGD} & \multirow{5}{*}{Linear} & 0.003                   &         & 0.5619                        \\ 
                       &                        &                         & 0.01                    &        & 0.5785                        \\ 
                       &                        &                         & 0.03                    &         & 0.5478                        \\ 
                       &                        &                         & 0.1                     &       & 0.4375                        \\ 
                       &                        &                         & 0.3                     &           & 0.2678                        \\ 
\hline
\multirow{3}{*}{a=0.1} & \multirow{3}{*}{M SGD} & \multirow{3}{*}{No}     & 0.003                   & 0.2629                        & 0.4276                        \\ 
                       &                        &                         & 0.01                    & 0.2847                        & 0.2959                        \\ 
                       &                        &                         & 0.03                    & 0.2754                        & 0.04848                       \\ 
\hline
\multirow{3}{*}{a=0.5} & \multirow{3}{*}{M SGD} & \multirow{3}{*}{No}     & 0.003                   & 0.3411                        & 0.3714                        \\ 
                       &                        &                         & 0.01                    & 0.426                         & 0.4691                        \\ 
                       &                        &                         & 0.03                    & 0.4418                        & 0.4412                        \\ 
\hline
\multirow{3}{*}{a=100} & \multirow{3}{*}{M SGD} & \multirow{3}{*}{No}     & 0.003                   & 0.3594                        & 0.5203                        \\ 
                       &                        &                         & 0.01                    & 0.4507                        & 0.4046                        \\ 
                       &                        &                         & 0.03                    & 0.4764                        & 0.5193                        \\
    \bottomrule
    \end{tabular}
    \end{threeparttable}
\caption{Summary of test accuracy on CIFAR-100 with MobileNet-V3.}
\label{tab:summary_of_cifar100_mobilenet}
\end{table}

\begin{figure*}[ht]
\small
\vspace{-0.2cm}
   \centering
   \!\!\!\!\!\!\!\!\subfigure[EfficientNet on GLD-23K]{\includegraphics[width=0.255\textwidth]{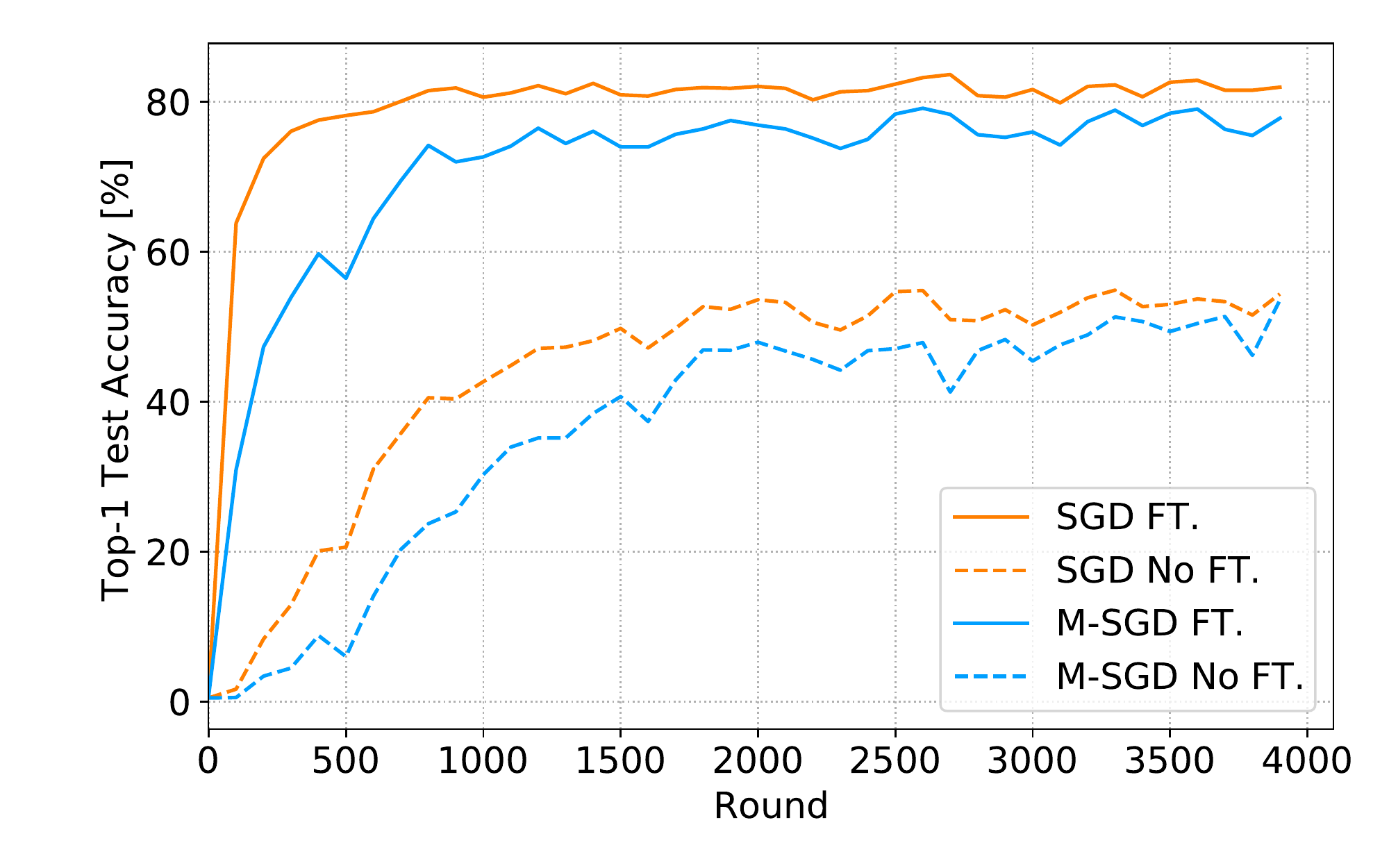}}\!\!\!\!
   \subfigure[MobileNet on GLD-23K]{\includegraphics[width=0.255\textwidth]{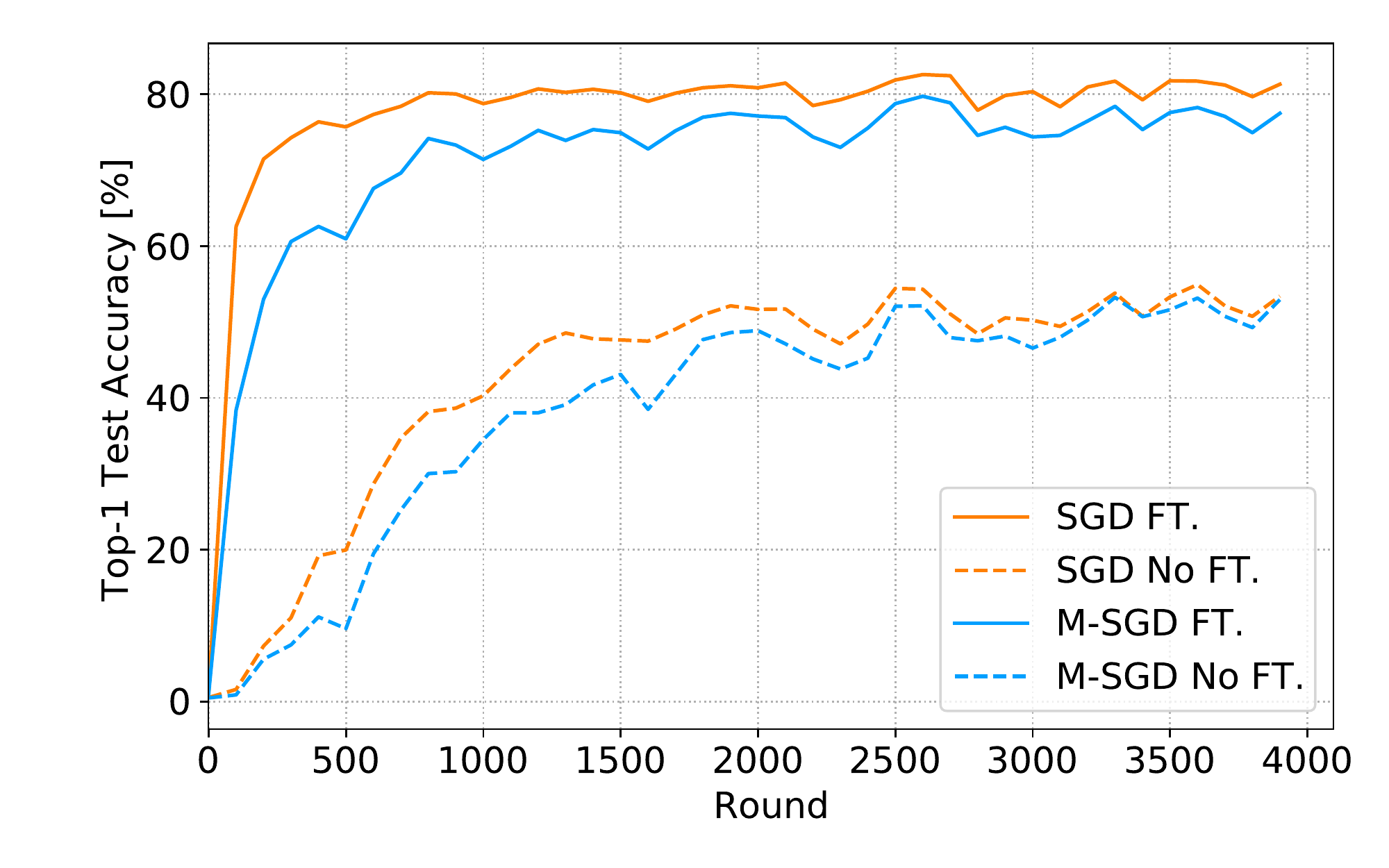}}\!\!\!\!
   \subfigure[EfficientNet on GLD-23K]{\includegraphics[width=0.255\textwidth]{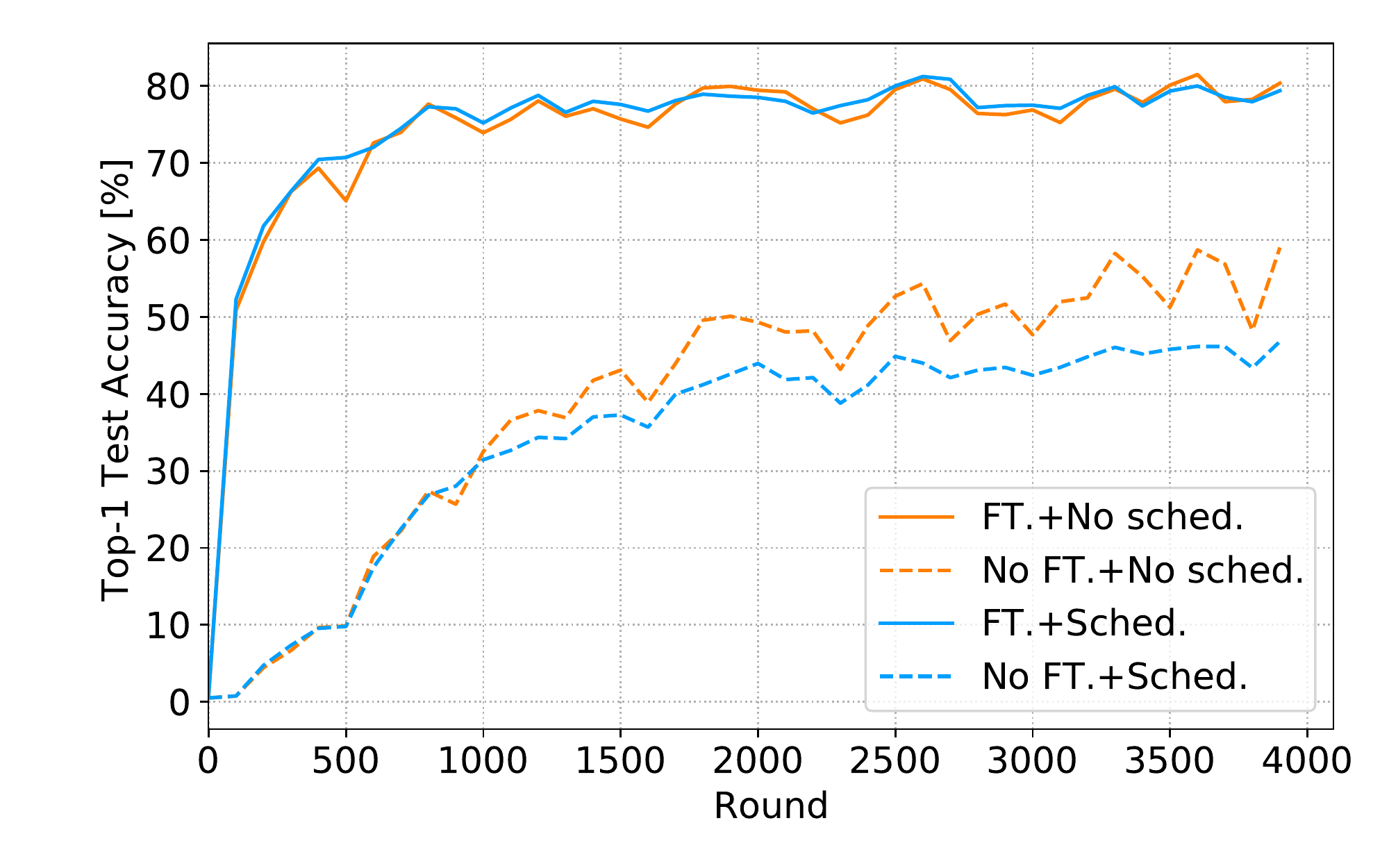}}\!\!\!\!
   \subfigure[MobileNet on GLD-23K]{\includegraphics[width=0.255\textwidth]{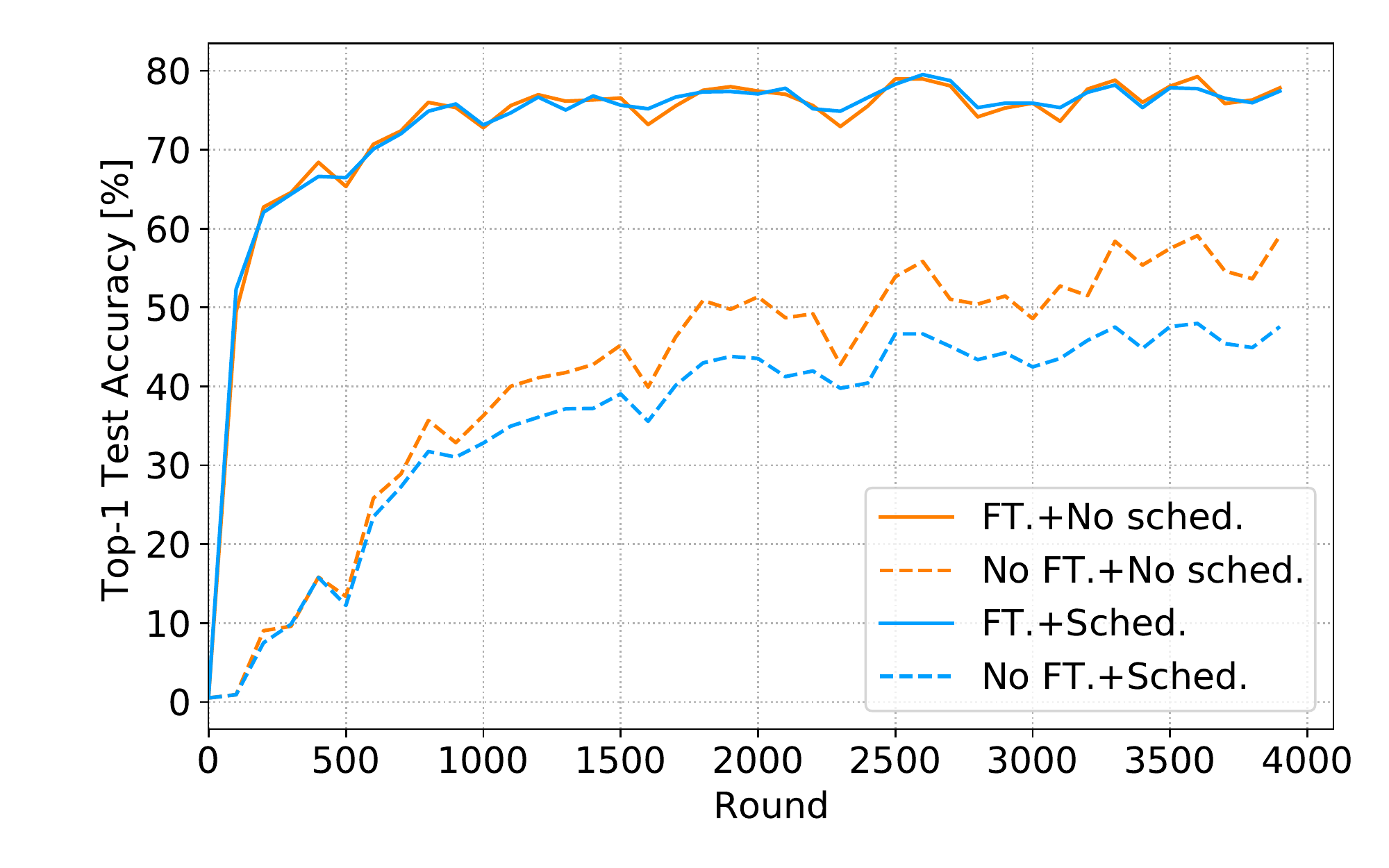}}\!\!\!\!\!
    \caption{Experiments on classification with different deep learning models and datasets. 
    Figure (a): Test accuracy of FedAvg with EfficientNet on GLD-23K, using or not using fine tuning, SGD or momentum SGD. Hyper-parameters of this figure can be found in table \ref{tab:summary_of_gld23k_efficientnet}.
    Figure (b): Test accuracy of FedAvg with MobileNet on GLD-23K, using or not using fine tuning, SGD or momentum SGD. Hyper-parameters of this figure can be found in table \ref{tab:summary_of_gld23k_mobilenet}.
    Figure (c): Test accuracy of FedAvg with EfficientNet on GLD-23K, using and not using fine tuning, learning rate sceduler or not. Hyper-parameters of this figure can be found in table \ref{tab:summary_of_gld23k_efficientnet}.
    Figure (d): Test accuracy of FedAvg with MobileNet on GLD-23K, using or not using fine tuning, learning rate sceduler or not. Hyper-parameters of this figure can be found in table \ref{tab:summary_of_gld23k_efficientnet}.}
    \label{fig:appendix_classification}
\end{figure*}

\begin{table}[!htp]
  \centering
  \fontsize{9}{9}\selectfont
  \begin{threeparttable}
    \begin{tabular}{cccccc}
    \toprule
Data                    & Opt.                   & Sched.                  & LR & SFL. Acc & FT. Acc  \\ 
\hline
\multirow{4}{*}{Cent.}  & \multirow{4}{*}{M SGD} & \multirow{4}{*}{Linear} & 0.03                    &          & 0.879                         \\ 
  
                        &                        &                         & 0.1                     &          & 0.8821                        \\ 
  
                        &                        &                         & 0.3                     &          & 0.8826                        \\ 
  
                        &                        &                         & 0.6                     &          & 0.88                          \\ 
\hline
\multirow{8}{*}{NonIID} & \multirow{6}{*}{M SGD} & \multirow{3}{*}{No}     & 0.03                    & 0.5319                        & 0.7938                        \\ 
  
                        &                        &                         & 0.1                     & 0.5901                        & 0.8035                        \\ 
  
                        &                        &                         & 0.3                     & 0.5615                        & 0.756                         \\ 
\cline{3-6}
                        &                        & \multirow{3}{*}{Linear} & 0.03                    & 0.3706                        & 0.7693                        \\ 
  
                        &                        &                         & 0.1                     & 0.4681                        & 0.7938                        \\ 
  
                        &                        &                         & 0.3                     & 0.5355                        & 0.7779                        \\ 
\cline{2-6}
                        & \multirow{2}{*}{SGD}   & \multirow{2}{*}{Linear} & 0.01                    & 0.01582                       & 0.5707                        \\ 
  
                        &                        &                         & 0.3                     & 0.5436                        & 0.8193                        \\
    \bottomrule
    \end{tabular}
    \end{threeparttable}
\caption{Summary of test accuracy on GLD-23K with EfficientNet-b0.}
\label{tab:summary_of_gld23k_efficientnet}
\end{table}

\begin{table}[!htp]
  \centering
  \fontsize{9}{9}\selectfont
  \begin{threeparttable}
    \begin{tabular}{cccccc}
    \toprule
Data                    & Opt.                   & Sched.                  & LR & SFL. Acc & FT. Acc  \\ 
\hline
\multirow{4}{*}{Cent.}  & \multirow{4}{*}{M SGD} & \multirow{4}{*}{Linear} & 0.03                    &          & 0.8698                        \\ 
  
                        &                        &                         & 0.1                     &          & 0.88                          \\ 
  
                        &                        &                         & 0.3                     &          & 0.8851                        \\ 
  
                        &                        &                         & 0.6                     &          & 0.8729                        \\ 
\hline
\multirow{8}{*}{NonIID} & \multirow{6}{*}{M SGD} & \multirow{3}{*}{No}     & 0.03                    & 0.4992                        & 0.7841                        \\ 
  
                        &                        &                         & 0.1                     & 0.5911                        & 0.7785                        \\ 
  
                        &                        &                         & 0.3                     & 0.4788                        & NaN      \\ 
\cline{3-6}
                        &                        & \multirow{3}{*}{Linear} & 0.03                    & 0.3267                        & 0.7601                        \\ 
  
                        &                        &                         & 0.1                     & 0.4758                        & 0.7744                        \\ 
  
                        &                        &                         & 0.3                     & 0.5294                        & 0.7749                        \\ 
\cline{2-6}
                        & \multirow{2}{*}{SGD}   & \multirow{2}{*}{Linear} & 0.01                    & 0.01327                       & 0.6085                        \\ 
  
                        &                        &                         & 0.3                     & 0.5339                        & 0.8132                        \\
    \bottomrule
    \end{tabular}
    \end{threeparttable}
\caption{Summary of test accuracy on GLD-23K with MobileNet-V3.}
\label{tab:summary_of_gld23k_mobilenet}
\end{table}

\begin{table}[!htp]
  \centering
  \fontsize{9}{9}\selectfont
  \begin{threeparttable}
    \begin{tabular}{cccccc}
    \toprule
Data                    & Opt.                   & Sched.                  & LR & FT. Acc  \\ 
\hline
\multirow{3}{*}{Cent.}  & \multirow{3}{*}{M SGD} & \multirow{3}{*}{Linear} & 0.003                   & 0.5967                        \\ 
\cline{4-5}
                        &                        &                         & 0.01                    & 0.7259                        \\ 
\cline{4-5}
                        &                        &                         & 0.03                    & 0.7565                        \\ 
\hline
\multirow{7}{*}{NonIID} & \multirow{7}{*}{M SGD} & \multirow{5}{*}{No}     & 0.003                   & 0.6554                        \\ 
\cline{4-5}
                        &                        &                         & 0.01                    & 0.7386                        \\ 
\cline{4-5}
                        &                        &                         & 0.03                    & 0.7611                        \\ 
\cline{4-5}
                        &                        &                         & 0.1                     & 0.7586                        \\ 
\cline{4-5}
                        &                        &                         & 0.3                     & 0.7489                        \\ 
\cline{3-5}
                        &                        & \multirow{2}{*}{Linear} & 0.03                    & 0.4957                        \\ 
\cline{4-5}
                        &                        &                         & 0.1                     & 0.7458                        \\
    \bottomrule
    \end{tabular}
    \end{threeparttable}
\caption{Summary of test accuracy on GLD-23K with ViT. Because the scratch training without pretrained model cannot get converge, we just ignore the results of scratch training here.}
\label{tab:summary_of_gld23k_vit}
\end{table}

As shown in figure \ref{fig:Image_classification_cifar100_convergence_mobilenet}, on CIFAR-100 dataset, the higher $\alpha$ makes training more difficult. The fine-tuning can increase the performance a lot.

Figure \ref{fig:Image_classification}(d) (In main paper) and \ref{fig:appendix_classification}(a)(b) compare the effect of SGD and Momentum SGD. On CIFAR-100 dataset, when $\alpha = 0.5$ and with fine-tuning, Momentum SGD has similar performance with SGD. However, on the GLD-23K dataset, the performance of Momentum SGD is worse than SGD on both fine-tuning and without fine-tuning. One potential reason is that the accumulated gradients make local clients go too far in its local direction, which means that the diversity of the model parameters between different clients is increased by Momentum SGD. Here the local gradient accumulation is much different from the centralized training, in which the accumulated gradient has information of many data samples. But in federated learning, each client only has accumulated gradients with information of its own data samples. Maybe this could be addressed if we can develop some protocols to preserve the gradient information of other clients.

Figure \ref{fig:Image_classification_cifar100_scheduler_efficientnet} (In main paper) and \ref{fig:appendix_classification}(c)(d) compare the effect of linear learning rate decay scheduler. On the GLD-23K dataset, when using fine-tuning, the performance between using the scheduler and not using is similar. However, when not using fine-tuning, using a scheduler cannot improve performance. But on CIFAR-100 dataset, when $\alpha=0.1$ and not using fine-tuning, we can see the benefit from the scheduler. It would be a non-trivial problem of how to use the learning rate scheduler in FedAvg. And another interesting result of FedAvg is that the test accuracy may drop after some training iterations. The reason for this may be the diverse optimal model parameter of different clients. Maybe some new learning rate scheduler could solve this problem well.

\subsubsection{Key Takeaways}

\textbf{Fine Tuning.} From experiment results, it is obvious that fine-tuning can improve the performance of FedAvg a lot.

\textbf{Optimizer.} The experiment results show that in FedAvg, Momentum SGD cannot guarantee better performance than SGD. Because each client has non-I.I.D. dataset, the direction of gradients have high diversity. The accumulated gradients may be in the wrong direction of the current global model, making the model parameters deviate from the right way. So in federated learning, those optimizers that need an accumulation of gradients, like momentum SGD and RMSprop, may not be suitable in federated learning. Maybe we can develop more new optimizers for FedAvg.

\textbf{Learning Rate Scheduler.} Experiment results show that the performance of learning rate decay is various. FedAvg has much more global epochs than traditional training, which makes the learning rate decay a lot in late training stages. Thus, it is necessary to consider a suitable learning rate scheduler for federated learning.

\textbf{Image Transform.} For federated learning, it can not be assumed that clients know the data distribution of other clients. In such a case, how to decide on a suitable Image Transform is not a trivial problem. Moreover, this is not limited to computer vision but also happens in other areas under federated learning settings. Because it is hard to get the global information of the data, we cannot directly do normalization of them.

\textbf{Data Augmentation.}
In recent years, there is a lot of advanced data augmentation techniques in computer vision proposed like AutoAugment \cite{autoaugment}, and GAN \cite{GAN}, and are exploited in many advanced models like EfficientNet \cite{efficientnet}. However, in federated learning, the whole data distribution cannot be attained by the server or any client. In this case, AutoAugment and GAN cannot be directly used in federated learning. This may lead to a performance drop.

\begin{figure}[htp]
    \centering
    \includegraphics[width=8.5cm]{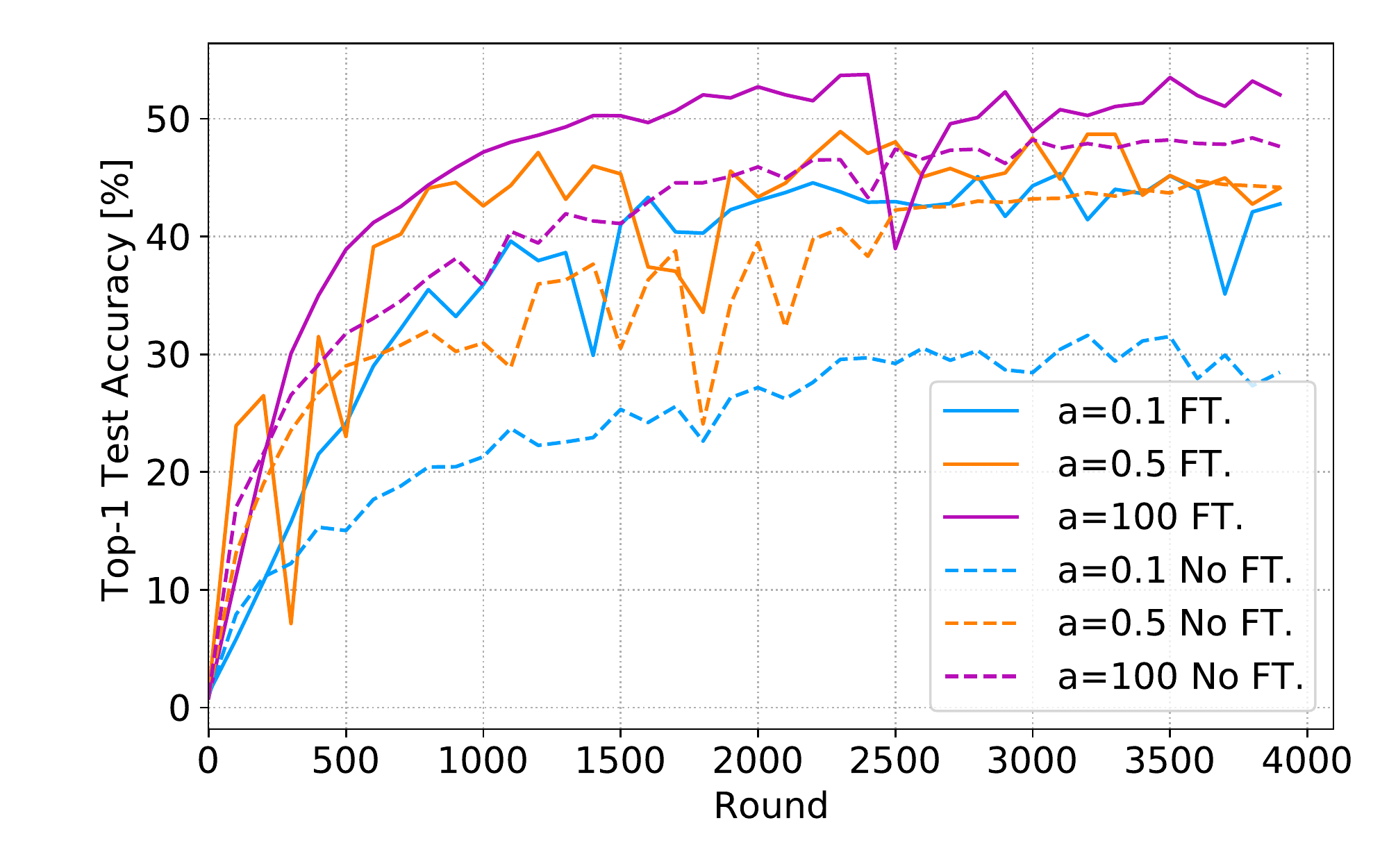}
    \caption{Test accuracy of FedAvg with MobileNet on CIFAR-100 with different Non-IID degree. Hyper-parameters of this figure can be found in table \ref{tab:summary_of_cifar100_mobilenet}}
    \label{fig:Image_classification_cifar100_convergence_mobilenet}
\end{figure}

\newpage
\clearpage

\begin{table*}[!htb]
\small
\centering
\caption{various datasets and models used in latest publications from the machine learning community}
\resizebox{\linewidth}{!}{
    \begin{threeparttable}
    \begin{tabular}{cp{0.6\columnwidth}cccc}\\
    \toprule
      \textbf{Conference} & \textbf{Paper Title}  &   \textbf{dataset}  & 
      \textbf{partition method}  &
      \textbf{model} &
      \textbf{worker/device} \\
       & & & & &\textbf{number} \\
        \midrule
      \multirow{2}{*}{ICML 2019} & 
      \multirow{2}{0.6\columnwidth}{Analyzing Federated Learning through an Adversarial Lens \cite{bhagoji2019analyzing}}&
      Fashion-MNIST & natural non-IID  &3 layer CNNs & 10\\
      \cmidrule(lr){3-6}
      & & UCI Adult Census datase &   -&fully connected neural network & 10\\
      
       \midrule
      \multirow{4}{*}{ICML 2019} & 
      \multirow{4}{0.6\columnwidth}{Agnostic Federated Learning \cite{mohri2019agnostic}}&
      UCI Adult Census datase & -  &logistic regression & 10\\
      \cmidrule(lr){3-6}
      & & Fashion-MNIST & -  &logistic regression & 10\\
      \cmidrule(lr){3-6}
      & & Cornell movie dataset &-   &two-layer LSTM mode & 10\\
      \cmidrule(lr){3-6}
      & & Penn TreeBank (PTB) dataset & - &two-layer LSTM mode & 10\\
      \midrule
      \multirow{2}{*}{ICML 2019} & 
      \multirow{2}{0.6\columnwidth}{Bayesian Nonparametric Federated Learning of Neural Networks \cite{yurochkin2019bayesian}}&
      MNIST & Dir(0.5)  &
      1 hidden layer neural networks & 10\\
      \cmidrule(lr){3-6}
      & & CIFAR10 & Dir(0.5) &
      1 hidden layer neural networks & 10\\
      \midrule
      \multirow{6}{*}{ICML 2020} & \multirow{6}{0.6\columnwidth}{
      Adaptive Federated Optimization \cite{reddi2020adaptive} } &
      CIFAR-100 & Pachinko Allocation Method &
      ResNet-18  &  10\\
      \cmidrule(lr){3-6}
     &  & FEMNIST &natural non-IID & CNN (2xconv) & 10\\
     \cmidrule(lr){3-6}
     &  & FEMNIST& natural non-IID  & Auto Encoder & 10\\
     \cmidrule(lr){3-6}
     &  & Shakespeare & natural non-IID  & RNN & 10 \\
     \cmidrule(lr){3-6}
     &  & StackOverflow & natural non-IID  & logistic regression& 10 \\
     \cmidrule(lr){3-6}
    &  & StackOverflow & natural non-IID  &1 RNN LSTM & 10 \\
    \midrule
      \multirow{3}{*}{ICML 2020} & 
      \multirow{3}{0.6\columnwidth}{FetchSGD: Communication-Efficient Federated Learning with Sketching \cite{rothchild_fetchsgd_2020}}&
      CIFAR-10/100 & 1 class / 1 client &
      ResNet-9 &- \\
      \cmidrule(lr){3-6}
      &  & FEMNIST & natural non-IID  & ResNet-101& - \\
      \cmidrule(lr){3-6}
     &  & PersonaChat & natural non-IID  & GPT2-small &-\\
     \midrule
      \multirow{5}{*}{ICML 2020} & 
      \multirow{5}{0.6\columnwidth}{Federated Learning with Only Positive Labels \cite{yu2020federated}} &
      CIFAR-10 & 1 class / client  &
      ResNet-8/32 & -\\
      \cmidrule(lr){3-6}
       & & CIFAR-100 & 1 class / client   & ResNet-56& - \\
       \cmidrule(lr){3-6}
      & & AmazonCAT &  1 class / client  & Fully Connected Nets & -\\
      \cmidrule(lr){3-6}
      & & WikiLSHTC &  1 class / client   & - &- \\
      \cmidrule(lr){3-6}
      & & Amazon670K &   1 class / client & - & -\\
      \midrule
      \multirow{2}{*}{ICML 2020} & 
      \multirow{2}{0.6\columnwidth}{SCAFFOLD: Stochastic Controlled Averaging for Federated Learning\cite{karimireddy2019scaffold}}&
      \multirow{2}{*}{EMNIST} & \multirow{2}{*}{ 1 class / 1 client}  &\multirow{2}{*}{
      Fully connected network} & \multirow{2}{*}{-}\\
      & & & & & \\
      & & & & & \\
      
      \midrule
      \multirow{1}{*}{ICML 2020} & 
      \multirow{1}{0.6\columnwidth}{From Local SGD to Local Fixed-Point Methods for Federated Learning \cite{malinovsky2020local}}& a9a(LIBSVM) & - & Logistic Regression &- \\
      \cmidrule(lr){3-6}
       & & a9a(LIBSVM) & - & Logistic Regression &- \\
       \midrule
      \multirow{4}{*}{ICML 2020} & 
      \multirow{4}{0.6\columnwidth}{Acceleration for Compressed Gradient Descent in Distributed and Federated Optimization \cite{li2020acceleration}}&
      a5a &  - & logistic regression & -\\
      \cmidrule(lr){3-6}
      & & mushrooms & - & logistic regression &- \\
      \cmidrule(lr){3-6}
      & & a9a & - & logistic regression & -\\
      \cmidrule(lr){3-6}
      & & w6a LIBSVM &-  & logistic regression & -\\
      \midrule
      \multirow{2}{*}{ICLR 2020} & 
      \multirow{2}{0.6\columnwidth}{Federated Learning with Matched Averaging \cite{wang2020federated}} &
      CIFAR-10 & - &
      VGG-9 & 16\\
      \cmidrule(lr){3-6}
      & & Shakespheare & sampling 66 clients  & 1-layer LSTM &66 \\
      \midrule
      \multirow{4}{*}{ICLR 2020} & 
      \multirow{4}{0.6\columnwidth}{Fair Resource Allocation in Federated Learning \cite{li2019fair}}&
      Synthetic dataset use LR & natural non-IID  &multinomial logistic regression & 10\\
\cmidrule(lr){3-6}
      & & Vehicle & natural non-IID  &SVM for binary classification & 10\\
      \cmidrule(lr){3-6}
      & & Shakespeare & natural non-IID  &RNN & 10\\
      \cmidrule(lr){3-6}
      & & Sent140 & natural non-IID  &RNN & 10\\
      \midrule
      \multirow{2}{*}{ICLR 2020} & 
      \multirow{2}{0.6\columnwidth}{On the Convergence of FedAvg on Non-IID Data \cite{li2019convergence}}&
      MNIST & natural non-IID  &logistic regression & 10\\
      \cmidrule(lr){3-6}
      & & Synthetic dataset use LR & natural non-IID  &logistic regression & 10\\
      
      \midrule
      \multirow{4}{*}{ICLR 2020} &
      \multirow{4}{0.6\columnwidth}{DBA: Distributed Backdoor Attacks against Federated Learning \cite{xie2019dba}}&
      Lending Club Loan Data & -  &3 FC & 10\\
      \cmidrule(lr){3-6}
      & & MNIST & -  &2 conv and 2 fc & 10\\
      \cmidrule(lr){3-6}
      & & CIFAR-10  & - &lightweight Resnet-18 & 10\\
      \cmidrule(lr){3-6}
      & & Tiny-imagenet & -  &Resnet-18 & 10\\
      \midrule
      \multirow{4}{*}{MLSys2020} & 
      \multirow{4}{0.6\columnwidth}{Federated Optimization in Heterogeneous Networks \cite{Sahu2018OnTC}} &
      MNIST & natural non-IID & multinomial logistic regression & 10\\
      \cmidrule(lr){3-6}
      & & FEMNIST & natural non-IID  &multinomial logistic regression & 10\\
      \cmidrule(lr){3-6}
      & & Shakespeare & natural non-IID  &RNN & 10\\
      \cmidrule(lr){3-6}
      & & Sent140 & natural non-IID  &RNN & 10\\
      \bottomrule
    \end{tabular}
    \begin{tablenotes}[para,flushleft]
      \footnotesize
      \item *Note: we will update this list once new publications are released.
    \end{tablenotes}
    \end{threeparttable}}
\label{tab:stat-datasets}
\end{table*}

\end{document}